\documentclass[runningheads]{llncs}
\usepackage{graphicx}

\usepackage{tikz}
\usepackage{comment}
\usepackage{amsmath,amssymb} 
\usepackage{color}
\usepackage{booktabs}
\usepackage{float}
\usepackage[normalem]{ulem}
\usepackage{multirow}
\usepackage{colortbl}
\usepackage{array}
\usepackage{adjustbox}
\newcommand*{\affaddr}[1]{#1} 
\newcommand*{\affmark}[1][*]{\textsuperscript{#1}}

\usepackage[accsupp]{axessibility}  
\definecolor{Teal}{RGB}{0,105,190}
\usepackage[pagebackref,breaklinks,colorlinks,citecolor=Teal]{hyperref}

\usepackage{wrapfig}
\usepackage{CJKutf8}
\usepackage{CJKutf8}

\begin{document}
\pagestyle{headings}
\mainmatter
\def\ECCVSubNumber{296}  

\title{BEAT: A Large-Scale Semantic and Emotional Multi-Modal Dataset for Conversational Gestures Synthesis} 

\titlerunning{BEAT}

\author{%
Haiyang Liu\affmark[1], Zihao Zhu\affmark[2], Naoya Iwamoto\affmark[3], Yichen Peng\affmark[4], \\
Zhengqing Li\affmark[3], You Zhou\affmark[3], Elif Bozkurt\affmark[5], Bo Zheng\affmark[3]}

\institute{\affaddr{\affmark[1]The University of Tokyo. \affmark[2]Keio University.\\
\affmark[2]Digital Human Lab, Huawei Technologies Japan K.K. \\
\affmark[4]Japan Advanced Institute of Science and Technology. \affmark[5]Huawei Turkey R\&D Center.}}
\authorrunning{H. Liu et al.}

\maketitle

\begin{figure}[h]
    \centering
    \vspace{-0.5cm}
    \includegraphics[trim=0 0 0 0, clip,width=0.9\textwidth]{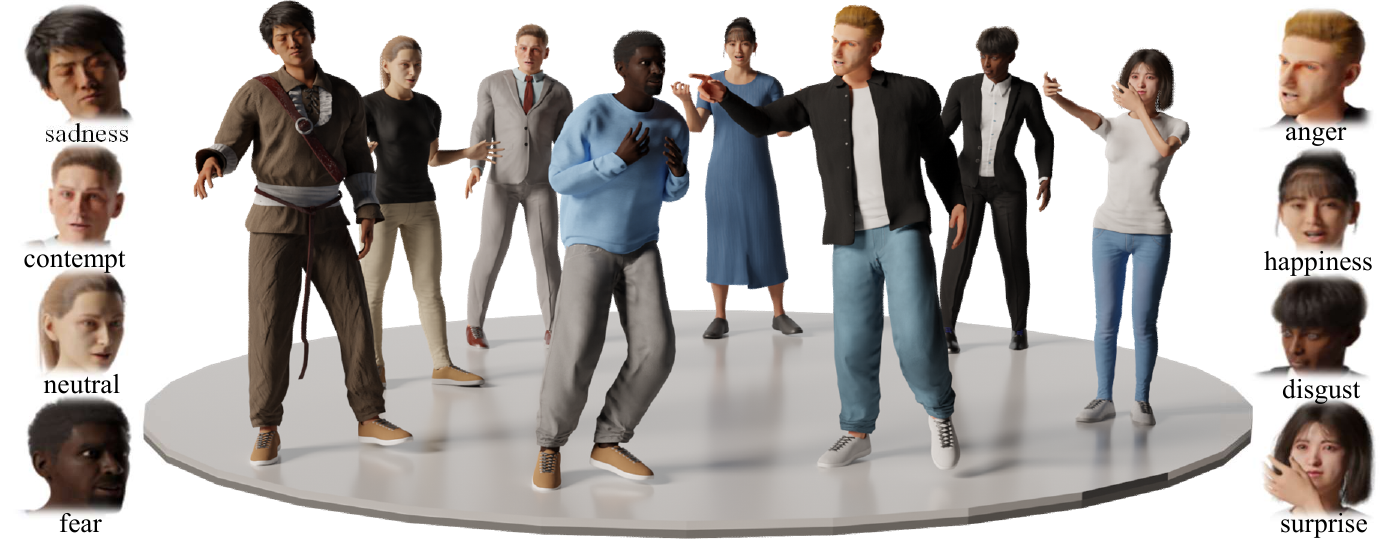}
    \vspace{-0.3cm}
   \caption{\textbf{Overview.} BEAT is a large-scale, multi-modal mo-cap human gestures dataset with semantic, emotional annotations, diverse speakers and multiple languages.}
    \label{concept}
    \vspace{-1.2cm}
\end{figure}

\begin{abstract}
Achieving realistic, vivid, and human-like synthesized conversational gestures conditioned on multi-modal data is still an unsolved problem due to the lack of available datasets, models and standard evaluation metrics. 
To address this, we build \textbf{B}ody-\textbf{E}xpression-\textbf{A}udio-\textbf{T}ext dataset, \textbf{BEAT}, which has i) 76 hours, high-quality, multi-modal data captured from 30 speakers talking with eight different emotions and in four different languages, ii) 32 millions frame-level emotion and semantic relevance annotations.
Our statistical analysis on BEAT demonstrates the correlation of conversational gestures with \textit{facial expressions}, \textit{emotions}, and \textit{semantics}, in addition to the known correlation with \textit{audio}, \textit{text}, and \textit{speaker identity}.
Based on this observation, we propose a baseline model, \textbf{Ca}scaded \textbf{M}otion \textbf{N}etwork \textbf{(CaMN)}, which consists of above six modalities modeled in a cascaded architecture for gesture synthesis. To evaluate the semantic relevancy, we introduce a metric, Semantic Relevance Gesture Recall (\textbf{SRGR}). 
Qualitative and quantitative experiments demonstrate metrics' validness, ground truth data quality, and baseline's state-of-the-art performance. 
To the best of our knowledge, BEAT is the largest motion capture dataset for investigating human gestures, which may contribute to a number of different research fields, including controllable gesture synthesis, cross-modality analysis, and emotional gesture recognition. The data, code and model are available on \url{https://pantomatrix.github.io/BEAT/}. 
\end{abstract}
\section{Introduction}
\label{sec:intro}

Synthesizing conversational gestures can be helpful for animation, entertainment, education and virtual reality applications. To accomplish this, the complex relationship between speech, facial expressions, emotions, speaker identity and semantic meaning of gestures has to be carefully considered in the design of the gesture synthesis models.  

While synthesizing conversational gestures based on audio \cite{li2021audio2gestures,yoon2020speech,ginosar2019learning} or text \cite{yoon2019robots,bhattacharya2021text2gestures,ali2020automatic,alexanderson2020generating} has been widely studied, synthesizing realistic, vivid, human-like conversational gestures is still unsolved and challenging for several reasons. 
i) \textbf{Quality and scale of the dataset}. 
Previously proposed methods \cite{yoon2020speech,li2021audio2gestures} were trained on limited mo-cap datasets \cite{takeuchi2017creating,ferstl2018investigating} or on pseudo-label \cite{ginosar2019learning,yoon2020speech,habibie2021learning} datasets (\textit{cf.} Table \ref{tab:tab1}), which results in limited generalization capability and lack of robustness.
ii) \textbf{Rich and paired multi-modal data}. Previous works adopted one or two modalities \cite{ginosar2019learning,yoon2019robots,yoon2020speech} to synthesize gestures and reported that conversational gestures are determined by multiple modalities together. However, due to the lack of paired multi-modal data, the analysis of other modalities, \textit{e.g.}, facial expression, for gesture synthesis is still missing. 
iii) \textbf{Speaker style disentanglement}. All available datasets, as shown in Table \ref{tab:tab1}, either have only a single speaker \cite{ferstl2018investigating}, or many speakers but different speakers talk about different topics \cite{habibie2021learning,yoon2020speech,ginosar2019learning}. Speaker-specific styles were not much investigated in previous studies due to the lack of data. 
iv) \textbf{Emotion annotation}. Existing work \cite{bhattacharya2021speech2affectivegestures} analyzes the emotion-conditioned gestures by extracting implicit sentiment features from texts. Due to the unlabeled, limited emotion categories in the dataset \cite{yoon2020speech}, it cannot cover enough emotion in daily conversations.  
v) \textbf{Semantic relevance}. Due to the lack of semantic relevance annotation, only a few works \cite{kucherenko2020gesticulator,yoon2020speech} analyze the correlation between generated gestures and semantics though listing subjective visualization examples. It will enable synthesizing context-related meaningful gestures if existing semantic labels of gestures.
In conclusion, the absence of a large-scale, high-quality multi-modal dataset with semantic and emotional annotation is the main obstacle to synthesizing human-like conversational gestures.

There are two design choices for collecting unlabeled multi-modal data, i) the pseudo-label approach \cite{ginosar2019learning,yoon2020speech,habibie2021learning}, \textit{i.e.}, extracting conversational gestures, facial landmark from in-the-wild videos using 3D pose estimation algorithms \cite{cao2019openpose} and ii) the motion capture approach \cite{ferstl2018investigating}, \textit{i.e.}, recording the data of speakers through predefined themes or texts. In contrast to the pseudo-labeling approach, which allows for low-cost, semi-automated access to large-scale training data, \textit{e.g.}, 97h \cite{yoon2020speech}, motion-captured data requires a higher cost and more manual work resulting in smaller dataset sizes, \textit{e.g.}, 4h \cite{ferstl2018investigating}. However, Due to the motion capture can be strictly controlled and designed in advance, it is able to ensure the quality and diversity of the data, e.g., eight different emotions of the same speaker, and different gestures of 30 speakers talking in the same sentences. Besides, high-quality motion capture data are indispensable to evaluate the effectiveness of pseudo-label training.

\begin{table}
\caption{\textbf{Comparison of Datasets.} We compare with all 3D conversational gesture and face datasets. ``\#", ``LM" and ``BSW" indicate the number, landmark and blendshape weight, respectively. \colorbox[rgb]{0.574,0.813,0.687}{best} and \colorbox[rgb]{0.855,0.933,0.894}{second} are highlighted. Our dataset is the largest mocap dataset with multi-modal data and annotations}
\begin{adjustbox}{width=\columnwidth, center}
\label{tab:tab1}
\begin{tabular}{l|c|cccccc|cc|cc}
        & \textbf{Quailty}                                                       & \multicolumn{6}{c|}{\textbf{Modality}}                                                                                                                                                                                                     & \multicolumn{2}{c|}{\textbf{Annotation}}                                  & \multicolumn{2}{c}{\textbf{Scale}}                                            \\
dataset &                                                                     & \#body                               & \#hand                               & face                                 & audio                                     & text                               & \#speaker                            & \#emo                                & sem                                & \#seq                                  & dura                                 \\ 
\hline
TED \cite{yoon2020speech}     & \multirow{2}{*}{\begin{tabular}[c]{@{}c@{}}pseudo\\label\end{tabular}} & 9                                    & -                                    & -                                    & En                                        &  \checkmark                                  & {\cellcolor[rgb]{0.574,0.813,0.687}} $>$100                                  & -                                    & -                                  & 1400                                   & {\cellcolor[rgb]{0.574,0.813,0.687}} 97h                                  \\
S2G \cite{ginosar2019learning,habibie2021learning}     &                                                                        & 14                                   & 42                                   & 2D LM                                   & En                                        & -                                  & 6                                    & -                                    & -                                  &  N/A                                      & 33h                                  \\ 
\hline
MPI \cite{volkova2014mpi}     & \multirow{5}{*}{mo-cap}                                                 & 23                                   & -                                    & -                                    & -                                         &  \checkmark                                  & 1                                   & {\cellcolor[rgb]{0.574,0.813,0.687}}11 & -                                  & 1408                                   & 1.5h                                 \\
VOCA \cite{cudeiro2019capture}    &                                                                        & -                                    & -                                    & {\cellcolor[rgb]{0.574,0.813,0.687}}3D Mesh                                   & En                                        & -                                  & 12                                   & -                                    & -                                  & 480                                    & 0.5h                                 \\
Takechi \cite{takeuchi2017creating} &                                                                        & 24                                   & 38                                   & -                                    & Jp                                        & -                                  & 2                                    & -                                    & -                                  & 1049                                   & 5h                                   \\
Trinity \cite{ferstl2018investigating} &                                                                        & 24                                   & 38                                   & -                                    & En                                        & \checkmark                                   & 1                                    & -                                    & -                                  & 23                                     & 4h                                   \\ 
\hline
BEAT (Ours)    & {\cellcolor[rgb]{0.574,0.813,0.687}}mo-cap                                & {\cellcolor[rgb]{0.574,0.813,0.687}}27 & {\cellcolor[rgb]{0.574,0.813,0.687}}48 & {\cellcolor[rgb]{0.855,0.933,0.894}}3D BSW & {\cellcolor[rgb]{0.574,0.813,0.687}}E/C/S/J & {\cellcolor[rgb]{0.574,0.813,0.687}}\checkmark & {\cellcolor[rgb]{0.855,0.933,0.894}}30 & {\cellcolor[rgb]{0.855,0.933,0.894}}8    & {\cellcolor[rgb]{0.574,0.813,0.687}}\checkmark & {\cellcolor[rgb]{0.574,0.813,0.687}}2508 & {\cellcolor[rgb]{0.855,0.933,0.894}}76h 
\end{tabular}
    \end{adjustbox}
\vspace{-0.6cm}
\end{table}

Based on the above analysis, to address these data-related problems, we built a mo-cap dataset \textbf{BEAT} containing semantic and eight different emotional annotations (\textit{cf.} Figure \ref{concept}), from 30 speakers in four modalities of \textbf{B}ody-\textbf{E}xpression-\textbf{A}udio-\textbf{T}ext, annotated in total of 30M frames. 
The motion capture environment is strictly controlled to ensure quality and diversity, with 76 hours and more than 2500 topic-segmented sequences. Speakers with different language mastery provided data in three other languages at different durations and in pairs. The ratio of actors/actresses, range of phonemes, and variety of languages are carefully designed to cover natural language characteristics. For emotional gestures, feedback on the speakers' expressions was provided by professional instructors during the recording process and re-recorded in case of non-expressive gesturing to ensure the expressiveness and quality of the entire dataset.
After statistical analysis on BEAT, we observed the correlation of conversational gestures with \textit{facial expressions}, \textit{emotions}, and \textit{semantics}, in addition to the known correlation with \textit{audio}, \textit{text}, and \textit{speaker identity}.  

Additionally, we propose a baseline neural network architecture,  \textbf{Ca}scaded \textbf{M}otion \textbf{N}etwork (\textbf{CaMN}), which learns synthesizing body and hand gestures by inputting all six modalities mentioned above. The proposed model consists of cascaded encoders and decoders for enhancing the contribution of audio and facial modalities. Besides, in order to evaluate the semantic relevancy, we propose \textbf{S}emantic-\textbf{R}elevant \textbf{G}esture \textbf{R}ecall (\textbf{SRGR}), which weights Probability of Correct Keypoint (PCK) based on  semantic scores of the ground truth data. Overall, our contributions can be summarized as follows:
\begin{itemize}
\setlength\itemsep{-.1em}
\item We release BEAT, which is the first gesture dataset with semantic and emotional annotation, and the largest motion capture dataset in terms of duration and available modalities to the best of our knowledge.

\item We propose CaMN as a baseline model that inputs audio, text, facial blendweight, speaker identity, emotion and semantic score to synthesize conversational body and hand gestures through cascaded network architecture.

\item  We introduce SRGR to evaluate the semantic relevancy as well as the human preference for conversational gestures. 

\end{itemize}
Finally, qualitative and quantitative experiments demonstrate the data quality of BEAT, the state-of-the-art performance of CaMN and the validness of SRGR.
\section{Related Work}
\label{sec:related}

\subsubsection{Conversational Gestures Dataset.} We first review mo-cap and pseudo-label conversational gestures datasets. Volkova \textit{et al.} \cite{volkova2014mpi} built a mo-cap emotional gestures dataset in 89 mins with text annotation, Takeuchi \textit{et al.} \cite{takeuchi2017speech} captured an interview-like audio-gesture dataset in total 3.5-hour with two Japanese speakers. Ferstl and Mcdonnell \cite{ferstl2018investigating} collected a 4-hour dataset, Trinity, with a single male speaker discussing hobbies, \textit{etc.}, which is the most common used mo-cap dataset for conversational gestures synthesis. On the other hand, Ginosar \textit{et al.} \cite{ginosar2019learning} used OpenPose \cite{cao2019openpose} to extract 2D poses from YouTube videos as training data for 144 hours, called S2G Dataset. Habibie \textit{et al.} \cite{habibie2021learning} extended it to a full 3D body with facial landmarks, and the last available data is 33 hours. Similarly, Yoon \textit{et al.} \cite{yoon2020speech} used VideoPose3D \cite{pavllo20193d} to build on the TED dataset, which is 97 hours with 9 joints on upper body. The limited data amount of mo-cap and noise in ground truth makes a trade-off for the trained network's generalization capability and quality. Similar to our work, several datasets are built for talking-face generation and the datasets can be divided into 3D scan face, \textit{e.g.}, VOCA \cite{takeuchi2017creating} and MeshTalk \cite{richard2021meshtalk} or RGB images \cite{alghamdi2018corpus,cao2014crema,cooke2006audio,jackson2014surrey,wang2020mead}. However, these datasets cannot be adopted to synthesize human gestures.

\vspace{-0.5cm}

\subsubsection{Semantic or Emotion-Aware Motion Synthesis.}
Semantic analysis of motion has been studied in the action recognition and the sign-language analysis/synthesis research domains. For example, in some of action recognition datasets \cite{ionescu2013human3,kay2017kinetics,carreira2017quo,chen2015utd,singh2010muhavi,bloom2012g3d,liu2016benchmarking,song2017end,perera2020multiviewpoint,wang2012robust} clips of action with the corresponding label of a single action, \textit{e.g.}, running, walking \cite{punnakkal2021babel} is used. Another example is audio-driven sign-language synthesis \cite{kapoor2021towards}, where hand gestures have specific semantics. However, these datasets do not apply to conversational gestures synthesis since gestures used in natural conversations are more complex than single actions, and their semantic meaning differs from sign-language semantics. Recently, Bhattacharya \cite{bhattacharya2021speech2affectivegestures}  extracted emotional cues from text and used them for gesture synthesis. However, the proposed method has limitations in the accuracy of the emotion classification algorithm and the diversity of emotion categories in the dataset.

\vspace{-0.5cm}

\subsubsection{Conditional Conversational Gestures Synthesis.} Early baseline models were released with datasets such as text-conditioned gesture \cite{yoon2019robots}, audio-conditioned gesture \cite{ginosar2019learning,takeuchi2017speech,ferstl2018investigating}, and audio-text-conditioned gesture \cite{yoon2020speech}. These baseline models were based on CNN and LSTM for end-to-end modelling. Several efforts try to improve the performance of the baseline model by input/output representation selection \cite{kucherenko2021moving,ferstl2021expressgesture}, adversarial training \cite{ferstl2020adversarial} and various types of generative modeling techniques \cite{wu2021probabilistic,lu2021double,wu2021modeling,ahuja2020style}, which can be summarized by "Estimating a better distribution of gestures based on the given conditions.". As an example, StyleGestures \cite{alexanderson2020style} uses Flow-based model \cite{henter2020moglow} and additional control signal to sample gesture from the distribution. Probabilistic gesture generation enables generating diversity based on noise, which is achieved by CGAN \cite{wu2021modeling}, WGAN \cite{wu2021probabilistic}. However, due to the lack of paired multi-modal data, the analysis of other modalities, \textit{e.g.}, facial expression, for gesture synthesis is still missing.  

\vspace{-0.5cm}

\section{BEAT: Body-Expression-Audio-Text Dataset}
\label{sec:other}
In this section, we introduce the proposed Body-Expression-Audio-Text (BEAT) Dataset. First, we describe the dataset acquisition process and then introduce text, emotion, and semantic relevance information annotation. Finally, we use BEAT to analyze the correlation between conversational gestures and emotions and show the distribution of semantic relevance.

\subsection{Data Acquisition}

\begin{figure}[h]
    \centering
    \includegraphics[trim=0 0 0 0, clip, width=0.9\textwidth]{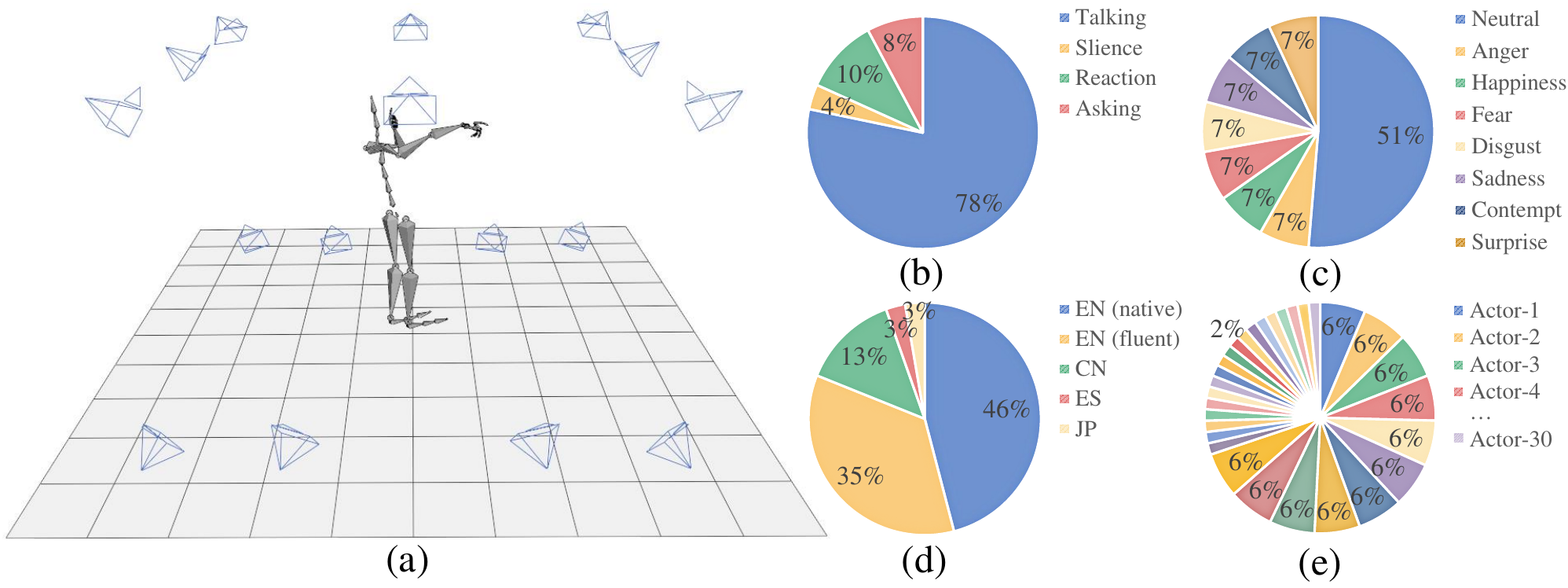}
    \vspace{-0.3cm}
    \caption{\textbf{Capture System and Subject Distribution of BEAT.} (a) A 16-camera motion capture system is adopted to record data in Conversation and Self-Talk sessions. (b) Gestures are divided into four categories in Conversation session. (c) Seven additional emotion categories are set in equal proportions in the self-talk session. Besides, (d) our dataset includes four languages which mainly consist of English, (e) by 30 speakers from ten countries with different recording duration.}
    \label{fig:fig2}
    \vspace{-1cm}
\end{figure}

\subsubsection{Motion Capture System.} The motion capture system shown in Figure \ref{fig:fig2}a, is based on 16 synchronized cameras recording motion at 120 Hz. We use Vicon's suits with 77 markers (\textit{cf.} supplementary materials for the location of markers on the body). The facial capture system uses ARKit with a depth camera on iPhone 12 Pro, which extracts 52 blendshape weights at 60 Hz. The blendshape targets are designed based on Facial Action Coding System (FACS) and are widely used by industry novice users. The audio is recorded in a 48KHz stereo.

\vspace{-0.5cm}

\subsubsection{Design Criteria.} BEAT is equally divided into \textit{conversation} and \textit{self-talk} sessions, which consist of 10-min and 1-min sequences, respectively. 
The conversation is between the speaker and the instructor remotely, \textit{i.e.}, to ensure only the speaker's voice is recorded. As shown in Figure \ref{fig:fig2}b, The speaker's gestures are divided into four categories talking, instantaneous reactions to questions, the state of thinking (silence) and asking. We timed each category's duration during the recording process. Topics were selected from 20 predefined topics, which cover 33\% and 67\% debate and description topics, respectively. \textit{Conversation} sessions would record the neutral conversations without acting to ensure the diversity of the dataset.
The \textit{self-talk} sessions consist of 120 1-minute self-talk recordings, where speakers answer questions about daily conversation topics, \textit{e.g.}, personal experiences or hobbies. The answers were written and proofread by three English native speakers, and the phonetic coverage was controlled to be similar to the frequently used 3000 words \cite{hornbyoxford}. We covered 8 emotions, \textit{neutral, anger, happiness, fear, disgust, sadness, contempt} and \textit{surprise}, in the dataset referring to \cite{livingstone2018ryerson} and the ratio of each emotion is shown in Figure \ref{fig:fig2}c. Among the 120 questions, 64 were for neutral emotions, and the remaining seven had eight questions each. Different speakers were asked to talk about the same content with their personalized gestures. Details about predefined answers and pronunciation distribution are available in the supplementary materials.

\vspace{-0.5cm}

\subsubsection{Speaker Selection and Language Ratio.} We strictly control the proportion of languages as well as accents to ensure the generalization capability of the dataset. As shown in Figure \ref{fig:fig2}d, the dataset consists mainly of English data: 60h (81\%), 12h of Chinese, 2h of Spanish and Japanese. The Spanish and Japanese are also 50\% of the size of the previous mo-cap dataset \cite{ferstl2018investigating}. The English component includes 34h of 10 native English speakers, including the US, UK, and Australia, and 26h of 20 fluent English speakers from other countries. As shown in Figure \ref{fig:fig2}e, 30 speakers (including 15 females) from different ethnicities can be grouped into two depending on their total recording duration as 4-hour (10 speakers) and 1-hour (20 speakers), where the 1-hour data is proposed for few-shot learning experiments. It is recommended to check the supplementary material for details of the speakers.

\vspace{-0.5cm}

\subsubsection{Recording.} Speakers were asked to read answers in self-talk sections proficiently. However, they were not guided to perform a specific style of gesture but were encouraged to show a natural, personal, daily style of conversational gestures. Speakers would watch 2-10 minutes of emotionally stimulating videos corresponding to different emotions before talking with the particular emotion. A professional speaker would instruct them to elicit the corresponding emotion correctly. We re-record any unqualified data to ensure the data's correctness and quality.

\subsection{Data Annotation}
\subsubsection{Text Alignment.} We use an in-house-built Automatic Speech Recognizer (ASR) to obtain the initial text for the conversation session and proofread it by annotators. Then, we adopt Montreal Forced Aligner (MFA) aligner \cite{mcauliffe2017montreal} for temporal alignment of the text with audio.

\vspace{-0.5cm}
\subsubsection{Emotion and Semantic Relevance.} The 8-class emotion label of self-talk is confirmed, and the on-site supervision guarantees the correctness. For the conversation session, annotators would watch the video with corresponding audio and gestures to perform frame-level annotation. For the semantic relevance, we get the score on a scale of 0-10 from assigned 600 annotators from Amazon Mechanical Turk (AMT). The annotators were asked to annotate a small amount of test data as a qualification check, of which only 118 annotators succeeded in the qualification phase for the final data annotation. We paid $\sim$ \$10 for each annotator per hour in this task.

\begin{figure}
    \centering
    \includegraphics[trim=0 0 0 0, clip, width=1\textwidth]{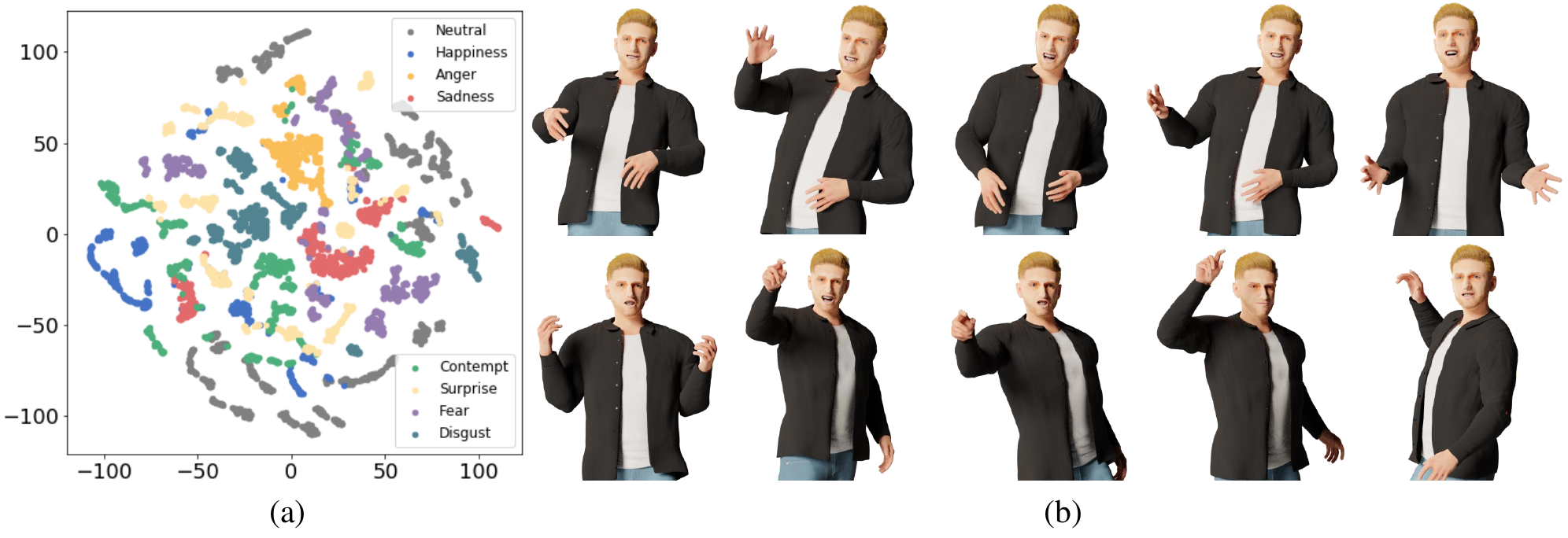}
    \vspace{-0.7cm}
    \caption{\textbf{Emotional Gesture Clustering and Examples.} (a) T-SNE visualization for gestures in eight emotion categories. Gestures with different emotions are basically distinguished into different groups, \textit{e.g.}, the Happiness (blue) and Anger (orange). (b) Examples of Happiness (top) and Anger gestures from speaker-2.}
    \label{fig:fig3}
\end{figure}

\subsection{Data Analysis}
The collection and annotation of BEAT have made it possible to analyze correlations between conversational gestures and other modalities. While the connection between gestures and audio, text and speaker identity has been widely studied. We further discuss the correlations between gestures, facial expressions, emotions, and semantics.

\vspace{-0.5cm}

\subsubsection{Facial Expression and Emotion.} 
Facial expressions and emotions were strongly correlated (excluding some of the lip movements), and we first analyze the correlation between conversational gestures and emotional categories here. As shown in Figure \ref{fig:fig3}a, We visualized the gestures in T-SNE based on a 2s-rotation representation, and the results showed that gestures have different characteristics in different emotions. For example, as shown in Figure \ref{fig:fig3}b, speaker-2 has different gesture styles when angry and happy, \textit{e.g.}, the gestures are larger and faster when angry. The T-SNE results also significantly differ between happy (blue) and angry (yellow). However, the gestures for the different emotions are still not perfectly separable by the rotation representation. Furthermore, the gestures of the different emotions appear to be confounded in each region, which is also consistent with subjective perceptions.

\vspace{-0.5cm}

\begin{figure*}[]
    \centering
    \includegraphics[trim=0 0 0 0, clip, width=0.95\textwidth]{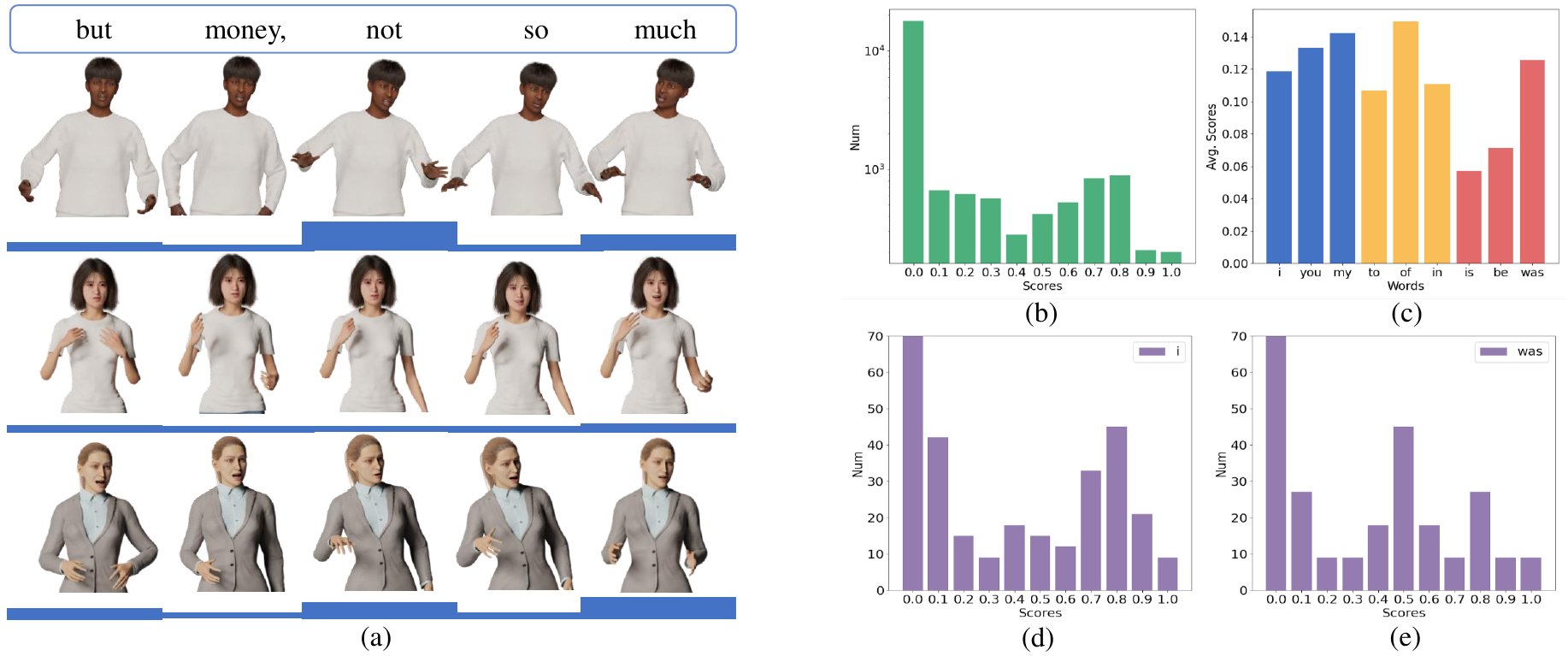}
    
    \caption{\textbf{Distribution of semantic labels.} (a) Different speaker ID speaks in a same phase happens different levels of semantic relevance and different styles of gesture. (b) The overall semantic distribution of BEAT. (c) The semantic relevance of the high frequency words which are grouped by their lexical in different color. (d, e) Different distribution of semantic relevance happens in words \textit{i} and \textit{was} even sharing almost the same level of semantic relevance.}
    \vspace{-0.6cm}
    \label{fig:fig4}
\end{figure*}

\subsubsection{Distribution of Semantic Relevance.} 
There is large randomness for the semantic relevance between gestures and texts, which is shown in Figure \ref{fig:fig4}, where the frequency, position and content of the semantic-related gestures vary from speaker to speaker when the same text content is uttered. 
In order to better understand the distribution of the semantic relevance of the gestures, we conducted a semantic relevance study based on four hours of two speakers' data. 
As shown in Figure ~\ref{fig:fig4}b, for the overall data, 83\% of the gestures have low semantic scores ($\leq$ 0.2). 
For the words-level, the semantic distribution varied between words, \textit{e.g.}, \textit{i} and \textit{was} which are sharing a similar semantic score but different in the score distribution.
Besides, Figure ~\ref{fig:fig4}c shows the average semantic scores of nine high-frequency words in the text corpus. It is to be mentioned that the scores of the \textit{Be-verbs} showed are comparatively lower than that \textit{Pronouns} and \textit{Prepositions} which are shown in blue and yellow, respectively.
Ultimately, it presents a different probability distribution to the semantically related gestures.

\section{Multi-Modal Conditioned Gestures Synthesis Baseline }
\label{sec:method}
\begin{figure*}[]
\begin{center}
\includegraphics[width=\columnwidth]{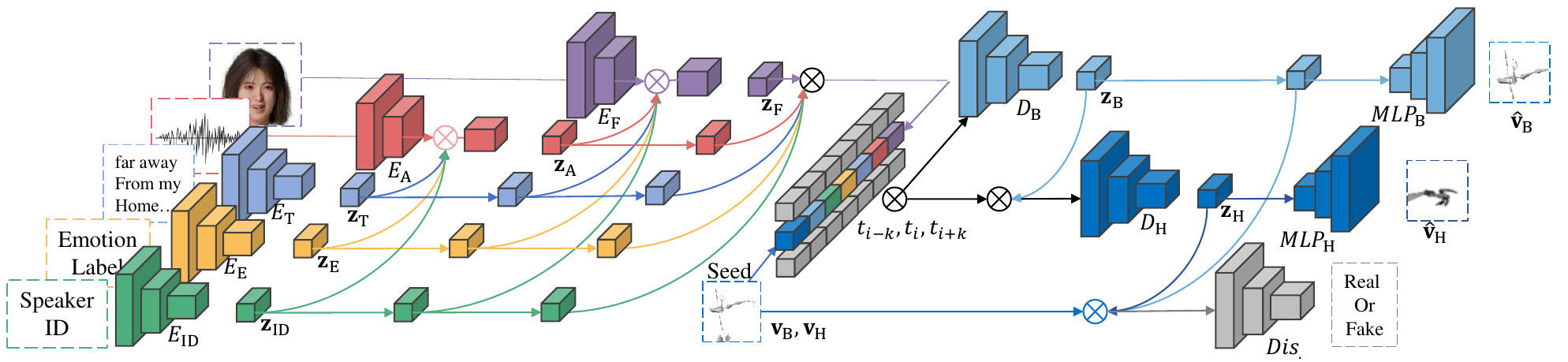}
\end{center}
\vspace{-0.5cm}
\caption{
%
%
\textbf{ Cascaded Motion Network (CaMN).} As a multi-modal gesture synthesis baseline, CaMN inputs \textit{text, emotion label, speaker ID, audio} and \textit{facial blendweight} in a cascaded architecture, the audio and facial feature will be extracted by concatenating the features of previous modalities. The fused feature will be reconstructed to body and hands gestures by two cascaded LSTM+MLP decoders.
}
\vspace{-0.5cm}
\label{fig:fig5}
\end{figure*}

In this section, we propose a baseline that inputs all the modalities for generating vivid, human-like conversational gestures. 
The proposed baseline, Cascaded Motion Network (CaMN),  is shown in Figure \ref{fig:fig5}, which encodes text, emotion condition, speaker identity, audio and facial blendshape weights to synthesize body and hands gestures in a multi-stage, cascade structure. In addition, semantic relevancy is adopted as a loss weight to make the network generate more semantic-relevant gestures. 
The text, audio and speaker ID encoders network selection are referred to \cite{yoon2020speech} and customized for better performance.
All input data have the same time resolution as the output gestures so that the synthesized gestures can be processed frame by frame through a sequential model. The gesture and facial blendshape weights are downsampled to 15 FPS, and the word sentence is inserted with padding tokens to correspond to the silence time in the audio.   

\vspace{-0.5cm}

\subsubsection{Text Encoder.} First, words are converted to word embedding set $\textbf{v}^\text{T} \in \mathbb{R}^{300} $ by pre-trained model in FastText \cite{bojanowski2017enriching} to reduce dimensions. Then, the word sets are fine-tuned by customized encoder $E_\text{T}$, which is a 8-layer temporal convolution network (TCN) \cite{bai2018empirical} with skip connections \cite{he2016deep}, as 
\begin{equation}
    z^\text{T}_{i} = E_\text{T}(v^\text{T}_{i-f}, ..., v^\text{T}_{i+f}), 
    \label{eq1}
\end{equation}
For each frame $i$, the TCN fusions the information from $2f=34$ frames to generate final latent feature of text, the set of features is note as $\textbf{z}^\text{T} \in \mathbb{R}^{128} $.

\vspace{-0.5cm}
\subsubsection{Speaker ID and Emotion Encoders.} The initial representation of speaker ID and emotion are both one-hot vectors, as $\textbf{v}^\text{ID} \in \mathbb{R}^{30}$ and  $\textbf{v}^\text{E} \in \mathbb{R}^{8}$. Follow the suggestion in \cite{yoon2020speech}, we use embedding-layer as speaker ID encoder, $E_\text{ID}$. As the speaker ID does not change instantly, we only use the current frame speaker ID to calculate its latent features. On the other hand, we use a combination of embedding-layer and 4-layer TCN as the emotion encoder, $E_\text{E}$, to extract the temporal emotion variations. 
\begin{equation}
    z^\text{ID}_{i} = E_\text{ID}(v^\text{ID}_{i}), z^\text{E}_{i} = E_\text{E}(v^\text{E}_{i-f}, ..., v^\text{E}_{i+f}), 
    \label{eq2}
\end{equation}
where $\textbf{z}^\text{ID} \in \mathbb{R}^{8}$ and  $\textbf{z}^\text{E} \in \mathbb{R}^{8} $ is the latent feature for speaker ID and emotion, respectively.

\vspace{-0.5cm}
\subsubsection{Audio Encoder.}
We adopt the raw wave representation of audio and downsample it to 16KHZ, considering audio as 15FPS, for each frame, we have $\textbf{v}^\text{A} \in \mathbb{R}^{1067}$. We feed the audio joint with the text, speakerID and emotion features into audio encoder $E_\text{A}$ to learn better audio features. As 
\begin{equation}
    z^\text{A}_{i} = E_\text{A}(v^\text{A}_{i-f}, ..., v^\text{E}_{i+f}; v^\text{T}_{i}; v^\text{E}_{i}; v^\text{ID}_{i}), 
    \label{eq4}
\end{equation} The $E_\text{A}$ consists of 12-layer TCN with skip connection and 2-layer MLP, features in other modifies are concatenated with the 12th layer audio features thus the final MLP layers are for audio feature refinement, and the final latent audio feature is $\textbf{z}^\text{A} \in \mathbb{R}^{128}$. 

\vspace{-0.5cm}
\subsubsection{Facial Expression Encoder.} We take the $\textbf{v}^\text{F} \in \mathbb{R}^{52}$ as initial representation of facial expression. 8-layer TCN and 2-layer MLP based encoder $E_\text{F}$ is adopt to extract facial latent feature $\textbf{z}^\text{F} \in \mathbb{R}^{32} $, as 
\begin{equation}
    z^\text{F}_{i} = E_\text{F}(v^\text{F}_{i-f}, ..., v^\text{F}_{i+f}; v^\text{T}_{i}; v^\text{E}_{i}; v^\text{ID}_{i}; v^\text{A}_{i}), 
    \label{eq5}
\end{equation}
the features are concatenated at 8th layer and the MLP is for refinement.

\vspace{-0.5cm}
\subsubsection{Body and Hands Decoders.} We implement the body and hands decoders in a separated, cascaded structure, which is based on \cite{ng2021body2hands} conclusion that the body gestures can be used to estimate hand gestures. These two decoders, $D_\text{B}$ and $D_\text{F}$ are based on the LSTM structure for latent feature extraction and 2-layer MLP for gesture reconstruction. They would combine the features of five modalities with previous gestures, \textit{i.e.}, seed pose, to synthesis latent gesture features $\textbf{z}^\text{B} \in \mathbb{R}^{256}$ and $\textbf{z}^\text{H} \in \mathbb{R}^{256}$. The final estimated body $\hat{\textbf{v}}^\text{B} \in \mathbb{R}^{27\times3}$ and hands $\hat{\textbf{v}}^\text{H} \in \mathbb{R}^{48\times3}$ are calculated as, 

\begin{equation}
    z^\text{M}_{i} = z^\text{T}_{i} \otimes z^\text{ID}_{i} \otimes z^\text{E}_{i} \otimes z^\text{A}_{i} \otimes z^\text{F}_{i} \otimes v^\text{B}_{i} \otimes v^\text{H}_{i},
    \label{eq6}
\end{equation} 
\begin{equation}
    \textbf{z}^\text{B} = D_\text{B}(z^\text{M}_{0}, ..., z^\text{M}_{n}), \textbf{z}^\text{H} = D_\text{H}(z^\text{M}_{0}, ..., z^\text{M}_{n}; \textbf{z}^\text{B}),
    \label{eq7}
\end{equation} 
\begin{equation}
    \hat{\textbf{v}}^\text{B} = MLP_{\text{B}}(\textbf{z}^\text{B}), 
    \hat{\textbf{v}}^\text{H} = MLP_{\text{H}}(\textbf{z}^\text{H}),
    \label{eq9}
\end{equation}     
$\textbf{z}^\text{M} \in \mathbb{R}^{549}$ is the merged features for all modalities. For Equation \ref{eq6}, the length for the seed pose is four frames.

\vspace{-0.5cm}
\subsubsection{Loss Functions.} The final supervision of our network is based on gesture reconstruction and the adversarial loss
\begin{equation}
\resizebox{.60\hsize}{!}{$
\ell_{\text{Gesture Rec.}} =\mathbb{E}\left[\left\|\textbf{v}^{B}-\hat{\textbf{v}}^{B}\right\|_{1}\right] + \alpha \mathbb{E}\left[\left\|\textbf{v}^{H}-\hat{\textbf{v}}^{H}\right\|_{1}\right],$}
\label{eq10}
\end{equation} 

\begin{equation}
\ell_\text{Adv.}=-\mathbb{E}[\log (Dis(\hat{\textbf{v}}^{B};\hat{\textbf{v}}^{H}))],
\end{equation} where the discriminator input to the adversarial training is only the gesture itself. We also adopt a weight $\alpha$ to balance the body and hands penalties. After that, during training, we adjust the weights of L1 loss, and adversarial loss using the semantic-relevancy label $\lambda$ The final loss function is \begin{equation}
\ell=\lambda \beta_{0} \ell_{\text {Gesture Rec.}}+\beta_{1} \ell_{\text {Adv}},
\label{eq12}
\end{equation} where $\beta_{0}$ and $\beta_{1}$ are predefined weight for L1 and adversarial loss. When semantic relevancy is high, we encourage the network to generate gestures spatially similar to ground truth as much as possible, thus strengthening the L1 penalty and decreasing the adversarial penalty. 
\section{Metric for Semantic Relevancy}
We propose the Semantic-Relevant Gesture Recall (SRGR) to evaluate the semantic relevancy of gestures, which can also be interpreted as whether the gestures are vivid and diverse. We utilize the semantic scores as a weight for the Probability of Correct Keypoint (PCK) between the generated gestures and the ground truth gestures. Where PCK is the number of joints successfully recalled against a specified threshold $\delta$. The SRGR metric can be calculated as follows:
\begin{equation}
\resizebox{.60\hsize}{!}{$
    D_{SRGR} = \lambda \sum \frac{1}{T \times J} \sum_{t=1}^{T} \sum_{j=1}^{J} \mathbf{1}\left[\left\|p_{t}^{j}-\hat{p}_{t}^{j}\right\|_{2}<\delta\right],$} 
\end{equation}
where $\mathbf{1}$ is the indicator function and $T, J$ is the set of frames and number of joints. 
We think the SRGR, which emphasizes recalling gestures in the clip of interest, is more in line with the subjective human perception of gesture's valid diversity than the L1 variance of synthesized gestures. 
\vspace{-0.5cm}

\section{Experiments}
In this section, we first evaluate the SRGR metric's validity, then demonstrate our dataset's data quality based on subjective experiments. Next, we demonstrate the validity of our baseline model using subjective and objective experiments, and finally, we discuss the contribution of each modality based on ablation experiments.

\subsection{Validness of SRGR}
\begin{figure*}[]

\begin{center}
\includegraphics[width=0.8\textwidth]{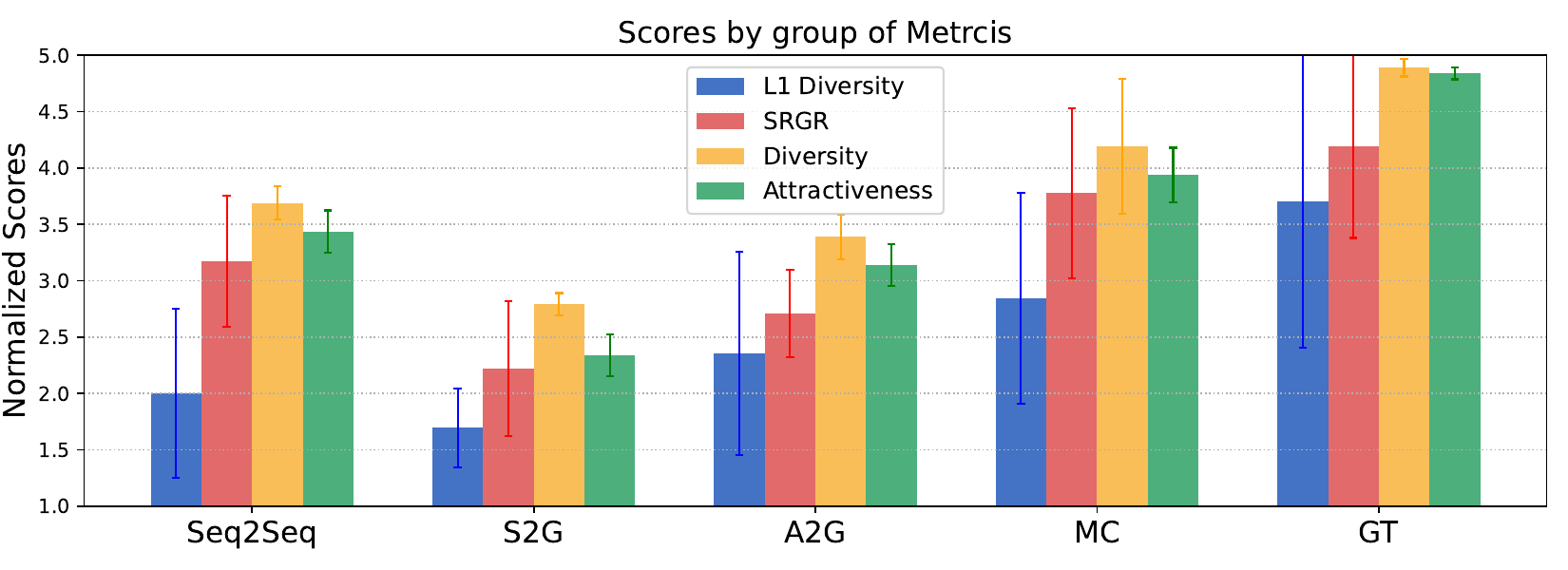}
\end{center}
\vspace{-0.5cm}
\caption{\textbf{Comparison of Metrics by Group.} SRGR shows the consistence with human perception, and lower variance comparing with L1 Diversity in evaluation.}
\vspace{-0.5cm}
\label{fig:fig6}
\end{figure*}


A user study is conducted to evaluate the validity of SRGR. 
Firstly, we randomly trim the motion sequences with rendered results into clips which are around 40s. 
For each clip, the participants are asked to evaluate the gesture based on its diversity which is the number of non-repeated gestures. 
Besides, the participants then need to score its attractiveness which should be based on the motion itself instead of the content of the speech.
Totally 160 participants took part in the evaluation study, and each of them evaluated 15 random clips of gestures. There are totally 200 gesture clips including the results generated by using the methods from Seq2Seq \cite{yoon2019robots}, S2G \cite{ginosar2019learning}, A2G \cite{li2021audio2gestures},  MultiContext \cite{yoon2020speech}, and ground truth, 40 clips for each with the same speaker data. Both of the questions follow a 5-points Likert scale. As shown in Figure \ref{fig:fig6}, we found a large variance in L1 diversity even though we used 100 gesture segments to calculate the average L1 distance, (usually around 40 segments \cite{li2021ai,li2021audio2gestures}). Secondly, generated results with strong semantic relevance but a smaller motion range, such as Seq2Seq, obtained a lower L1 diversity than A2G, which has a larger motion range, yet the statistical evidence that humans feel that Seq2Seq has higher diversity than A2G. An explanation is a human evaluating diversity not only on the range of motion but also on some other implicit features, such as expressiveness and semantic relevancy of the motion.


\subsection{Data quality}

To evaluate the captured ground truth motion data quality, we compare our proposed dataset with the widely used mocap dataset Trinity \cite{ferstl2018investigating} and in-the-wild dataset S2G-3D \cite{ginosar2019learning,habibie2021learning}. We conducted the user study by comparing clips sampled from ground truth and generated results using motion synthesis networks trained in each dataset. The Trinity dataset has a total of 23 sequences, with 10 minutes each. We randomly divide the data into 19:2:2 for train/valid/test since there is no standard for splitting.

\begin{table}
\vspace{-0.5cm}
\centering
\caption{\textbf{User Study Comparison with Trinity for Data Quality.} Comparing with Trinity \cite{ferstl2018investigating}, BEAT get \colorbox[rgb]{0.574,0.813,0.687}{higher} user preference score in terms of ground truth data quality. ``-b" and ``-h" indicate body and hands, respectively.}
\label{tab:tab2}
\resizebox{.90\hsize}{!}{
\begin{tabular}{cccccccccccccccc}
                            & \multicolumn{3}{c}{correctness-b}                                                                                        &  & \multicolumn{3}{c}{correctness-h}                                                                                        &  & \multicolumn{3}{c}{diversity~}                                                                                           &  & \multicolumn{3}{c}{synchrony}                                                                                             \\
                            & S2G                                    & A2G                                    & GT                                     &  & S2G                                    & A2G                                    & GT                                     &  & S2G                                    & A2G                                    & GT                                     &  & S2G                                    & A2G                                    & GT                                      \\ 
\hline
\multicolumn{1}{l}{Trinity \cite{ferstl2018investigating}} & 38.8                                   & 37.0                                   & 43.8                                   &  & 15.3                                   & 14.6                                   & 11.7                                   &  & 42.1                                   & 36.7                                   & 40.2                                   &  & 40.9                                   & 36.3                                   & 46.4                                    \\
\multicolumn{1}{l}{BEAT (Ours)}    & {\cellcolor[rgb]{0.574,0.813,0.687}}61.2 & {\cellcolor[rgb]{0.574,0.813,0.687}}63.0 & {\cellcolor[rgb]{0.574,0.813,0.687}}56.2 &  & {\cellcolor[rgb]{0.574,0.813,0.687}}84.7 & {\cellcolor[rgb]{0.574,0.813,0.687}}85.4 & {\cellcolor[rgb]{0.574,0.813,0.687}}88.3 &  & {\cellcolor[rgb]{0.574,0.813,0.687}}57.9 & {\cellcolor[rgb]{0.574,0.813,0.687}}63.3 & {\cellcolor[rgb]{0.574,0.813,0.687}}59.8 &  & {\cellcolor[rgb]{0.574,0.813,0.687}}59.1 & {\cellcolor[rgb]{0.574,0.813,0.687}}63.7 & {\cellcolor[rgb]{0.574,0.813,0.687}}53.6 
\end{tabular}}
\vspace{-0.5cm}
\end{table}

\vspace{-0.5cm}

\begin{table}[h]
\centering
\caption{\textbf{User Study Comparison with S2G-3D.} BEAT get similar user preferences in terms of naturalness. Based on the score, the model trained on BEAT dataset would be fitted into a more physically correct, diverse, and attractive distribution. }
\label{tab:tabs1}
\resizebox{.90\hsize}{!}{
\begin{tabular}{lclclclc}
                    & ~ naturalness                                    &  & correctness                                      &  & diversity                                        &  & attractiveness                                    \\ 
\hline
S2G-3D \cite{ginosar2019learning,habibie2021learning}              & 33.03 ± 1.93                                     &  & 21.17 ± 2.84                                     &  & 29.17 ± 1.81                                     &  & 28.79 ± 2.53                                      \\
BEAT (conversation) & {\cellcolor[rgb]{0.573,0.812,0.686}}34.16 ± 2.16 &  & {\cellcolor[rgb]{0.573,0.812,0.686}}39.94 ± 3.97 &  & 34.69 ± 1.76                                     &  & 29.90 ± 2.19                                      \\
BEAT (self-talk)    & 32.81 ±1.79                                      &  & 38.89 ± 3.75                                     &  & {\cellcolor[rgb]{0.573,0.812,0.686}}36.14 ± 1.99 &  & {\cellcolor[rgb]{0.573,0.812,0.686}}42.31 ± 2.40 
\end{tabular}
}
\end{table}
\vspace{-0.5cm}

We used S2G \cite{ginosar2019learning}, as well as the SoTA algorithm A2G \cite{li2021audio2gestures}, to cover both GAN and VAE models. The output layer of the S2G model was adapted for outputting 3D coordinates. In the ablation study, the final generated 3D skeleton results were rendered and composited with audio for comparison in the user study. A total of 120 participant subjects compared the clips randomly sampled from Trinity and our dataset, with 5-20s in length. The participants were asked to evaluate gestures correctness, \textit{i.e.}, physical correctness, diversity and gesture-audio synchrony. Furthermore, the body and hands were evaluated separately for the gesture correctness test. The results are shown in Table \ref{tab:tab2}, demonstrating that our dataset received higher user preference in all aspects. Especially for the hand movements, we outperformed the Trinity dataset by a large margin. This is probably due to the noise of the past motion capture devices and the lack of markers on the hands. Table \ref{tab:tabs1} shows preference ratios (\%) of 60 subjects who watch 20 random rendered 3D skeletons pairs per subjective test. Based on the score, the model trained on the BEAT dataset would be fitted into a more physically correct, diverse, and attractive distribution.

\subsection{Evaluation of the baseline model}
\begin{table}
\begin{minipage}{\textwidth}
\begin{minipage}[t]{0.50\textwidth}
\centering
\caption{\textbf{Evaluation on BEAT.} Our CaMN performs \colorbox[rgb]{0.574,0.813,0.687}{best} in the term of FGD, SRGR and BeatAlign, all methods are trained on our dataset (BEAT) }
\label{tab:tab3}
\resizebox{.99\hsize}{!}{
\begin{tabular}{lccc}
       & FGD $\downarrow$           & ~SRGR $\uparrow$ & BeatAlign $\uparrow$ \\ 
\hline
Seq2Seq \cite{yoon2019robots} & 261.3         &  0.173     & 0.729      \\
S2G \cite{ginosar2019learning}     & 256.7          & 0.092      &  0.751     \\
A2G \cite{li2021audio2gestures}     & 223.8          & 0.097      &  0.766     \\
MultiContext \cite{yoon2020speech}    & 176.2          & 0.196      &  0.776     \\
CaMN (Ours)  &{\cellcolor[rgb]{0.574,0.813,0.687}} 123.7 &{\cellcolor[rgb]{0.574,0.813,0.687}}  0.239     &{\cellcolor[rgb]{0.574,0.813,0.687}}  0.783    
\end{tabular}
}
\end{minipage}
\begin{minipage}[t]{0.50\textwidth}
\centering
\label{tab:tab4}
\caption{\textbf{Results of Ablation Study.}}
\resizebox{.99\hsize}{!}{
\begin{tabular}{lccc}
               & FGD $\downarrow$ & BGSR $\uparrow$ & BeatAlign  $\uparrow$ \\ 
\hline
full cascated  & 123.7    &   0.239      &    0.783        \\
w/o cascaded   & 137.2   &  0.207   &  0.776    \\  
w/o text       & 149.4    & \cellcolor[rgb]{0.935,0.660,0.660} 0.171   &   0.781             \\
w/o audio      & 155.7     & 0.225              &  \cellcolor[rgb]{0.935,0.660,0.660} 0.733               \\
w/o speaker ID &\cellcolor[rgb]{0.963,0.81,0.81} 159.1     &    0.207          &  0.774            \\
w/o face       & \cellcolor[rgb]{0.935,0.660,0.660}163.3     & 0.217              & \cellcolor[rgb]{0.963,0.81,0.81} 0.767              \\
w/o emotion    & 151.7     & 0.231              &  0.775              \\
w/o semantic   & 151.8     &\cellcolor[rgb]{0.963,0.81,0.81}   0.194          &   0.786               
\end{tabular}}
\end{minipage}
\end{minipage}
\end{table}
\subsubsection{Training Setting.} We use the Adam optimizer \cite{kingma2014adam} to train at a learning rate of 2e-4, and the 4-speaker data is trained in an NVIDIA V100 environment. For evaluation metrics, L1 has been demonstrated unsuitable for evaluating the gesture performance  \cite{li2021audio2gestures,yoon2020speech} thus, we adopt FGD \cite{yoon2020speech} to evaluate the generated gestures' distribution distance with ground truth. It computes the distance between latent features extracted by a pretrained network, we use an LSTM-based autoencoder as the pretrained network. In addition, we adopt SRGR and BeatAlign to evaluate diversity and synchrony. 
BeatAlign \cite{li2021ai} is a Chamfer Distance between audio and gesture beats to evaluate gesture-audio beat similarity.

\vspace{-0.5cm}
\subsubsection{Quantitative Results.} The final results are shown in Table \ref{tab:tab3}. In addition to S2G and A2G, we also compare our results with text-to-gesture and audio\&test-to-gesture algorithm, Seq2Seq \cite{yoon2019robots} and MultiContext \cite{yoon2020speech}. The results show that both our end2end model and cascaded model archive SoTA performance in all metrics (\textit{cf.} supplementary materials for video results). 

\subsection{Ablation Study.}
\subsubsection{Effectiveness of Cascaded Connection.} As shown in Table 5, in contrast to the end-to-end approach, the cascaded connection can achieve better performance because we introduce prior human knowledge to help the network extract features of different modalities. 

\subsubsection{Effectiveness of Each Modality.} We gradually removed the data of one modality during the experiment (\textit{cf.} Table 5). Synchrony would significantly be reduced after removing the audio, which is intuitive. However, it still maintains some synchronizations, such as the padding and time-align annotation of the text and the lip motion of the facial expression. In contrast, eliminating weighted semantic loss improves synchrony, which means that semantic gestures are usually not strongly aligned with audio perfectly. There is also a relationship between emotion and synchrony, but speaker ID only has little effect on synchrony. The removal of audio, emotion, and facial expression does not significantly affect the semantic relevant gesture recall, which depends mainly on the text and the speaker ID. Data from each modality contributed to improving the FGD, which means using different modalities of data enhances the network's mapping ability. The unities of audio and facial expressions, especially facial expressions, improve the FGD significantly. We found that removing emotion and speaker ID also impacts the FGD scores. This is because using the integrated network increases the diversity of features, which leads to a diversity of results, increasing the variance of the distribution and making it more like the original data.

\subsubsection{Emotional Gestures.} As shown in Table \ref{s2}, we train a classifier by an additional 1DCNN + LSTM network and invite 60 subjects each to classify 12 random real test clips (with audio). The classifier is trained and tested on speaker-4's ground truth data.

\begin{table}[h]
\caption{\textbf{Emotional Gesture Classification.} The classification accuracy (\%) gap between the test real and generated data (1344 clips, 10s each) is 15.85.}
\centering
\label{s2}
\resizebox{.90\hsize}{!}{
\begin{tabular}{lccccccccc}
          & neutral     & happiness     & sadness    & anger     & surprise    & contempt     & fear     & disgust     & avg.   \\ 
\hline
human & 84.29 & 74.86 & 82.65 & 88.36 & 76.12 & 71.59 & 80.94 & 72.33 & 78.89  \\
real & 51.06 & 98.68 & 85.08 & 38.78 & 99.39 & 81.08 & 99.95 & 99.62 & 83.26  \\
generated & 36.95 & 76.83 & 62.17 & 37.46 & 77.91 & 70.61 & 81.32 & 83.03 & 67.41 
\end{tabular}
}
\end{table}
\vspace{-0.5cm}

\subsection{Limitation}
\subsubsection{Impact of Acting.} Self-Talk sessions might reflect the impact of acting, which is inevitable and controlled. \textit{Inevitable:} The impact is probably caused by pre-defined content. However, to explore the semantic-relevancy and personality, it is necessary to control the variables, \textit{i.e.}, different speakers should talk in the same text and emotion so that the personality can be carefully explored. \textit{Controlled.} Speakers recorded the conversation session first and were encouraged to keep the same style as the conversation. We also filtered out about 21h of data and six speakers due to inconsistencies in their styles.

\subsubsection{Calculation of SRGR.} SRGR now is calculated based on semantic annotation, which has a limitation for an un-labelled dataset. To solve this problem, training a scoring network or semantic discriminator is a possible direction.
\section{Conclusion}
We build a large-scale, high-quality, multi-modal, semantic and emotional annotated dataset to generate more human-like, semantic and emotional relevant conversational gestures. Together with the dataset, we propose a cascade-based baseline model for gesture synthesis based on six modalities and achieve SoTA performance. Finally, we introduce SRGR for evaluating semantic relevancy. In the future, we plan to expand cross-data checks for AU and emotion recognition benchmarks. Our dataset and the related statistical experiments could benefit a number of different research fields, including controllable gesture synthesis, cross-modality analysis and emotional motion recognition in the future.

\section{Acknowledgements}
This work was conducted during Haiyang Liu, Zihao Zhu, and Yichen Peng’s internship at Tokyo Research Center. We thank Hailing Pi for communicating with the recording actors of the BEAT dataset.
\bibliographystyle{splncs04}
\bibliography{egbib}
\newpage
\appendix

\textbf{A.} Annotation Interface and Measurement of Agreement.

\textbf{B.} Details of Text Content and Speaker Information.

\textbf{C.} Details of Data Release Formats. 

\textbf{D.} Additional Discussions for SRGR, FGD and
BeatAlign. 

\textbf{E.} More Subjective Results and Videos. 

\textbf{F.} Details of Baselines Training Setting. 

\textbf{G.} Answers in Self-Talk Session (\textit{cf}. answers.doc).

\section{Annotation Interface and Measurement of Agreement.}
\begin{figure}[h]
    \centering
    \includegraphics[width=.90\textwidth]{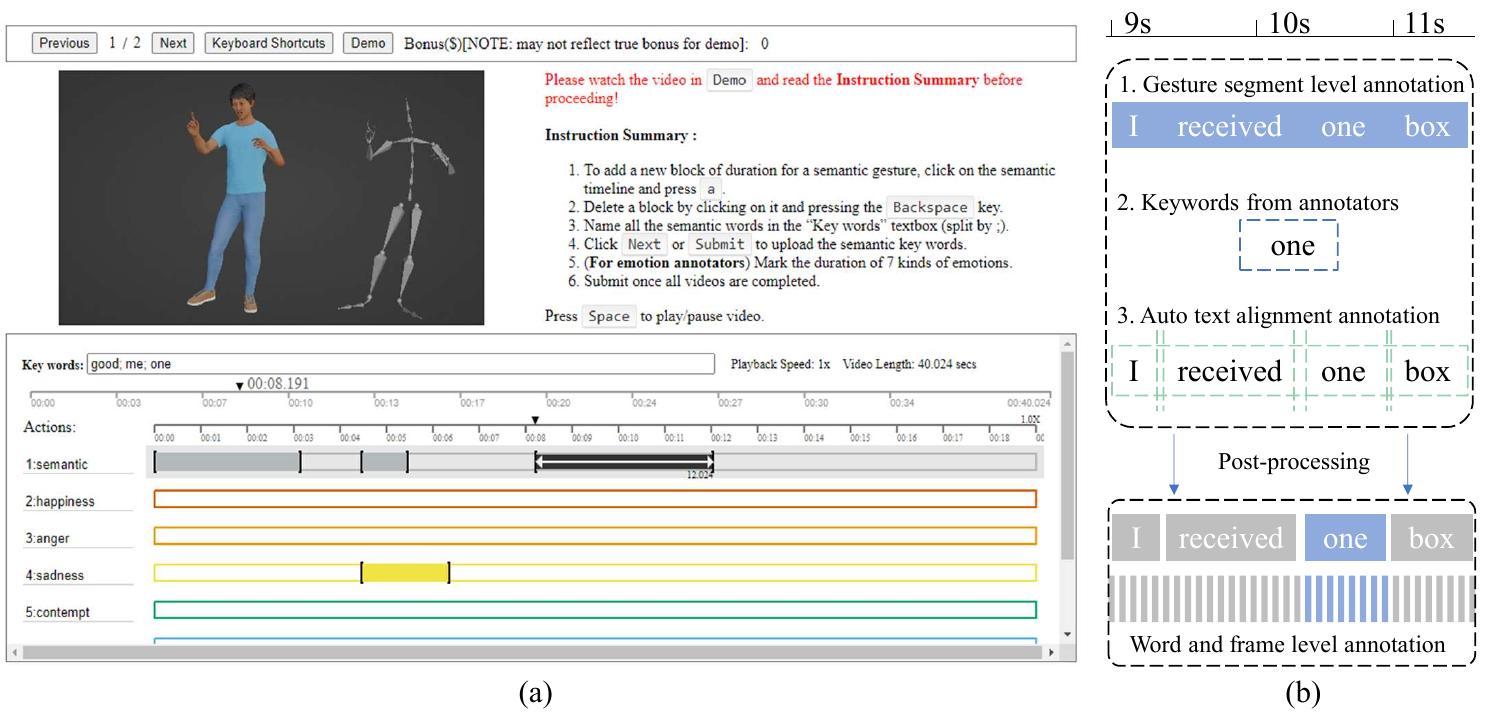}
    \caption{\textbf{Annotation Interface and Post-Processing.} (a) Annotators will give the gesture segment level and keyword level annotations. (b) Post-processing algorithm will generate a frame-level semantic relevance score using segment, keyword, and text alignment annotation.  }
    \label{fig:spfig1}
\end{figure}

Our annotation interface, as shown in Figure \ref{fig:spfig1}a, is adapted from a modified version of VGG Image Annotator (VIA) \cite{dutta2019vgg} by \cite{punnakkal2021babel}. We use the same interface for annotating both emotions and semantics. However, annotations are performed by a different group of annotators for each task. For emotion annotations, two annotators annotate the start and end times of each video segment of \textit{Conversation} sessions. For semantics annotations, annotators would i) agree or disagree whether the current gesture is semantically related to the text content according to their perception; ii) if agreed, they annotate the start and end times for the current gesture; iii) and then, the select keyword(s) that they think the gesture exactly corresponds to from the list of keywords separated by a comma.  

Our post-processing algorithm will input the separated keywords, text alignment and gesture segment-level semantic annotation to generate frame-level annotation. As shown in Figure \ref{fig:spfig1}b, the final semantic relevance for each frame is calculated as the multiplication of the gesture segment semantic score and the keyword semantic score. It is to be mentioned that two levels of semantic annotation can also be adopted separately.

We calculate the inter-rater agreement rate of the annotations by Measurement of inter-rater reliability for both emotional and semantic annotations. For the emotion annotations, we present scores in Table \ref{tab:tabsup1}, the agreement is marked only two annotators give the same label. The final agreement for around 16M frames is 96\%, which is high enough that we did not conduct the annotation for emotion with more than two annotators.

\begin{figure}
\caption{\textbf{Statistic on Emotion Annotation.} \textit{Left}: The sum of duration in emotion annotations (in seconds). \textit{Right}: Distribution of annotations number in each clip.}
\label{tab:tabsup1}
\begin{tabular}{c}
\begin{minipage}[b]{0.50\hsize}
\centering
\includegraphics[trim=20 5 10 00, clip, width=\hsize]{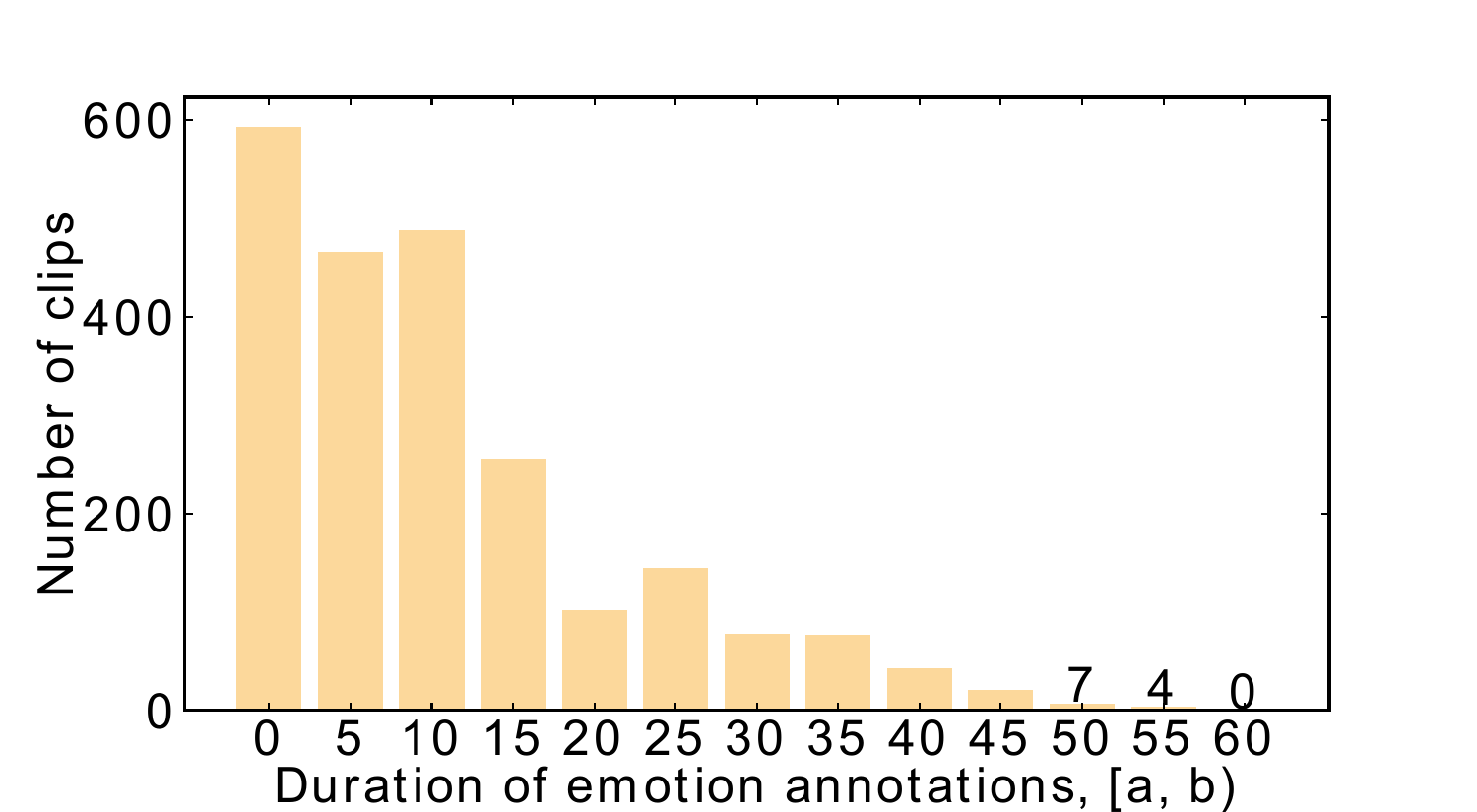}
\end{minipage}
\begin{minipage}[b]{0.50\hsize}
\centering

\includegraphics[trim=20 5 10 00, width=\hsize]{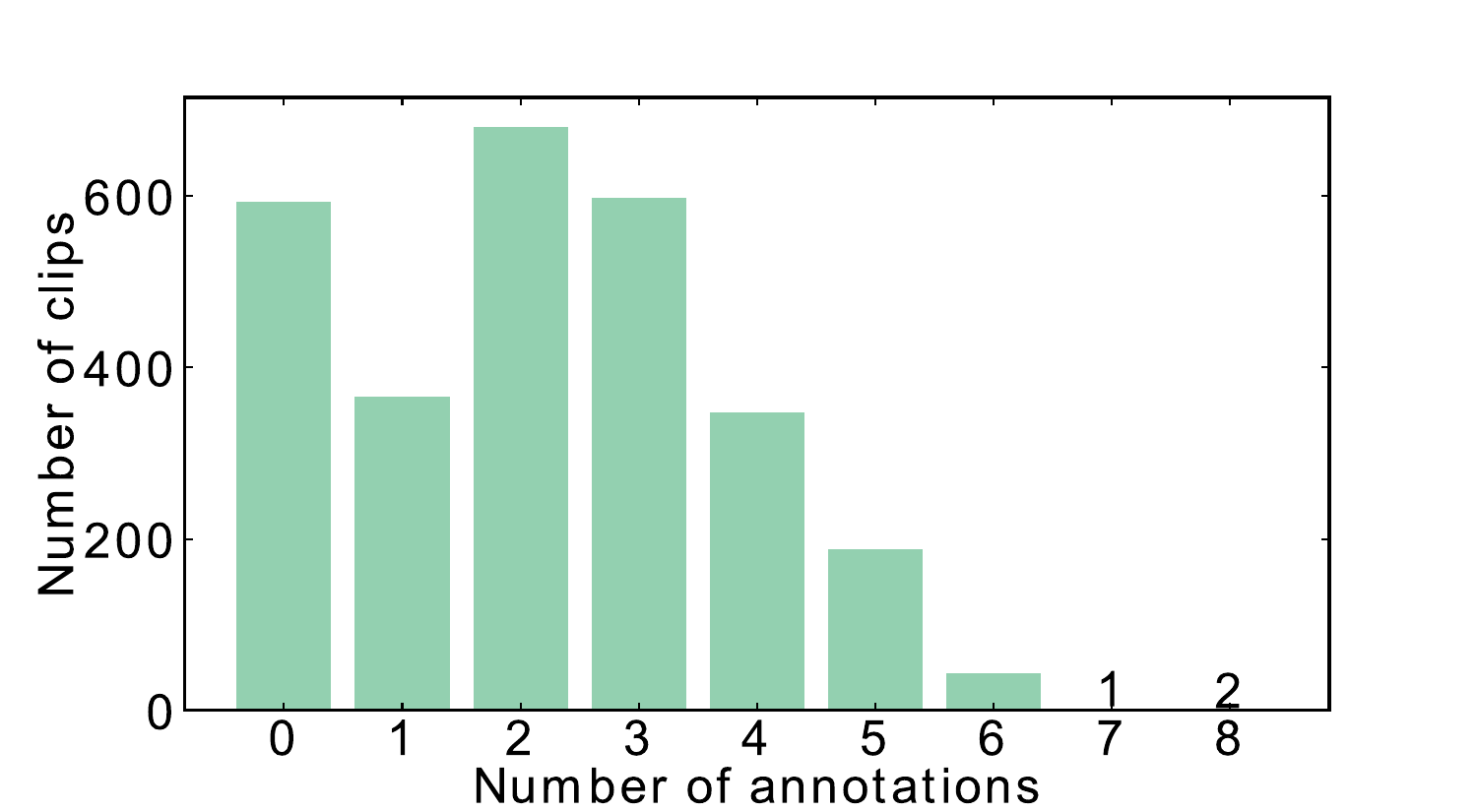}
\end{minipage}
\end{tabular}
\end{figure}

\begin{table}[h]
\centering
\caption{Example of inter-rater reliability calculation for emotion annotation. a1 and a2 indicates annotator 1 and 2, respectively.}
\label{tab:tabsup1}
\resizebox{.25\hsize}{!}{
\begin{tabular}{lccc}
frame & a1 & a2 & agree?  \\ 
\hline
01     & 0  & 0  & 1       \\
02     & 1  & 2  & 0       \\
03     & 1  & 1  & 1       \\
04     & 1  & 0  & 0       \\
\hline
avg.  &    &    & 0.5    
\end{tabular}}
\end{table}

\section{Details of Text Content and Speaker Information.}
\begin{figure}
    \centering
    \includegraphics[width=0.99\textwidth]{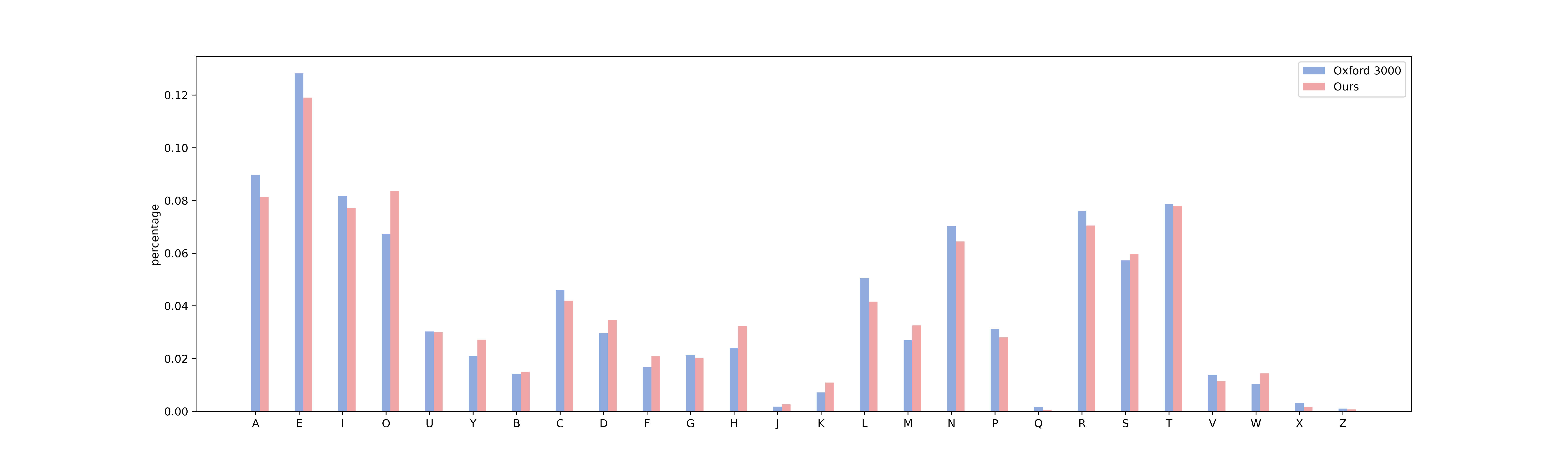}
    \vspace{-0.3cm}
    \caption{\textbf{Distribution of Vowel and Consonant.} The distribution of our corpus is basically consistent with that of frequently used 3000 words in \cite{hornbyoxford}.}
    \label{fig:spfig2}
    \vspace{-0.3cm}
\end{figure}

The distribution of vowels and consonants for the BEAT dataset is shown in Figure \ref{fig:spfig2}, which is basically consistent with the frequently used 3000 words \cite{hornbyoxford}. For the \textit{Conversation} sessions, the questions are selected from Table \ref{tab:sptab4}. There are ten topics for debate and introduction, respectively, which are related to daily conversation topics. For \textit{Self-Talk} session, the full answer list, including 120 answers, is attached at the end of this supplementary material, which includes the translation of four languages for 30 answers proofread by native speakers.  

\begin{table}
\centering
\caption{\textbf{List of Conversation Topics.} }
\label{tab:sptab4}
\resizebox{.95\hsize}{!}{$
\begin{tabular}{ll}
                    & Topics                                                                                              \\ 
\hline
1                   & Do you think smart-phone has destroyed communication among friends and family?                                    \\
\multirow{2}{*}{2}  & Now people usually work from home, for you, which one you prefer?~                    \\
                    & Face-to-face communication or work from home?        \\
3                   & In general, people are living longer now. Why? Discuss the causes of this phenomenon.  \\
\multirow{2}{*}{4}  & Some people believe that the Earth is being harmed (damaged) by human activity ~                            \\
                    & Others feel that human activity makes the Earth a better place to live. What is your opinion?.                                                         \\
\multirow{2}{*}{5}  & Some people are always in a hurry to go places and get things done then take a rest.                                 \\
                    & Other people prefer to take their time and live life at a slower pace. Which do you prefer?         \\
\multirow{2}{*}{6}  & Some people say that advertising encourages us to buy things we really do not need.                 \\
                    & Others say that advertisements tell us about new products that may improve our lives.~ ~            \\
\multirow{2}{*}{7}  & Some jobs (such as Salesmen) can be done by human or by robots (AI).~                       \\
                    & Which do you prefer? get the service from human or robots (AI)?                                  \\
\multirow{2}{*}{8}  & Television, news, and other media pay too much attention to the                                     \\
                    & personal lives of famous people such as public figures and celebrities                              \\
9                   & Is it more important to be able to work with a group of people on a team or to work independently?                \\
\multirow{2}{*}{10} & Do you agree or disagree with the following statement?~                                             \\
                    & Only people who earn a lot of money are successful.                                                 \\ 
\hline
11                  & Introduce some places, cities, countries, or even planets                                                                           \\
12                  & Introduce some celebrities, artists or you friends                                                                           \\
13                  & Introduce some sports or physic knowledge.                                                                                \\
14                  & Introduce some pets or plants.                                                                            \\
15                  & Introduce some electronic Products, cars or other vehicles                                                                                  \\
16                  & Introduce some video games, musics, books, TV programs, stories, or movies.                                                                                  \\
17                  & Introduce some foods or chemical phenomenon.                                                                     \\
18                  & Introduce some historical events.                                                                     \\
19                  & Introduce some psychological phenomenon.                                                                                 \\
20                  & Introduce some military doctrine.                                                                
\end{tabular}$}
\end{table}

\begin{table}
\centering
\caption{\textbf{Speaker information} in terms of gender, originating country, native English speaker, recording duration in English and in the second language, age and ethnicity. In the gender column, \textbf{M} and \textbf{F} stand for male and female speakers, respectively. The native column shows whether the speaker is a native English speaker. Duration is represented in hours (h).}
\label{tab:sptab3}
\begin{tabular}{ccccccccc}
   & gender & country & native & dura. & other lan. & other dura. & age & ethnicity  \\ 
\hline
1  & M   & US & \checkmark  & 4h       & -              &  -              & 25  & Caucasian  \\
2  & M   & US & \checkmark & 4h       & -              & -               & 32  & Caucasian      \\
3  & M   & US & \checkmark  & 4h       & -              & -               & 40  & African      \\
4  & M   & Australia & \checkmark   & 4h       & -              & -               & 26  & Asian       \\
5  & M   & UK & \checkmark   & 1h       & -              &-              & 30    & Caucasian      \\
6  & F & US     & \checkmark  & 4h       & -              & -               & 27  & Caucasian  \\
7  & F & US & \checkmark   & 4h       & -              & -               & 30  & Caucasian       \\
8  & F & US & \checkmark   & 4h       & -              & -               & 31  &  Asian \\
9  & F & UK & \checkmark   & 4h         & -              & -               & 32    & Caucasian    \\
10 & F & UK & \checkmark   & 1h       & -              & -               & 35    & Caucasian     \\
11 & M   & Arab   & -          &  4h          & -              & -              & 38    & African      \\
12 & M   & Thailand  & -          &  1h           & -              &  -              & 32    & Asian      \\
13 & M   & China   & -        &  1h           & Chinese        & 4h              & 25  & Asian      \\
14 & M   & China   & -        &  1h           & Chinese        & 1h              & 24  & Asian      \\
15 & M   & China   & -         & 1h            & Chinese        & 1h             & 40  & Asian      \\
16 & M   & China   & -         & 1h            & -           & -             & 32  & Asian      \\
17 & M   & Japan  & -         & 1h            & Japanese    & 1h              & 32    & Asian            \\
18 & M   & Japan  & -         & 1h            & -               & -               & 22    & Asian           \\
19 & M   & Peru    & -          & 1h            & Spanish        & 1h              & 27  & Caucasian  \\
20 & M   & Spain   & -          & 1h             & Spanish        & 1h              & 30  & Caucasian  \\
21 & F & China   & -          & 1h           & Chinese        & 4h              & 31  & Asian      \\
22 & F & China   & -         & 1h            & Chinese        & 1h              & 24  & Asian      \\
23 & F & China   & -        & 1h            & Chinese        & 1h              & 26    & Asian      \\
24 & F & China   & -          & 1h            & -       &  -          & 32  & Asian      \\
25 & F & Japan   & -          & 1h            & Japanese           &  1h              & 24    & Asian      \\
26 & F & Japan   & -          & 1h            & -              &  -              & 26 & Asian      \\
27 & F & Iran    & -           & 1h            & -              & -               & 31     & African           \\
28 & F & Jamaica    & -           & 4h            & -              & -               & 33    &  African           \\
29 & F & Jamaica    & -            & 1h            & -              & -               & 24    & African           \\
30 & F & Russia    & -           & 1h             & -              & -               & 25    & Caucasian          
\end{tabular}
\end{table}

Holding the motivation for investigating the style differences among speakers, we collected data from speakers of various countries, gender, ages and ethnicity. We then modelled these differences with explicit controls on styles. During data collection, we made sure the actor style was consistent. We filtered out about 21 hours of data and six speakers due to inconsistencies in their styles. The actor presented significantly different gestures in \textit{Self-Talk} and \textit{Conversation} sessions. For example, some speakers gestured a lot during the conversation sessions but demonstrated almost no gestures in the self-talk sessions. The number of effective speakers is 30, and the corresponding recorded data duration is 76 hours. Speaker information and duration of their recordings are available in Table \ref{tab:sptab3}. We have 34 and 26 hours of recordings from native and fluent English speakers, respectively, and the native/fluent duration ratio is 1.307.

\section{Details of Data Release Format.}
Our final released data file format is listed below:

i) Motion capture data in BVH file format for body and hand gestures.

ii) Audio recordings in stereo WAV file format.

iii) Facial expression blendshape weights in JSON file format.

iv) Facial mesh data in FBX file format for 8 speakers.

v) Text-audio alignment annotation data in TextGrid file format.

vi) Semantics and emotion annotations in text file format.

For body and hand gestures, the position of the markers are shown in Figure \ref{fig:spfig2} (image from\footnote{\url{https://sketchfab.com/3d-models})} and joint names are shown in Figure \ref{fig:spfig3}, respectively. We use the rotation information in Euler angles as the motion representation, \textit{i.e.}, 75$\times$3 rotations + 1$\times$3 root  translation. 

\begin{figure}
    \centering
    \includegraphics[width=.7\textwidth]{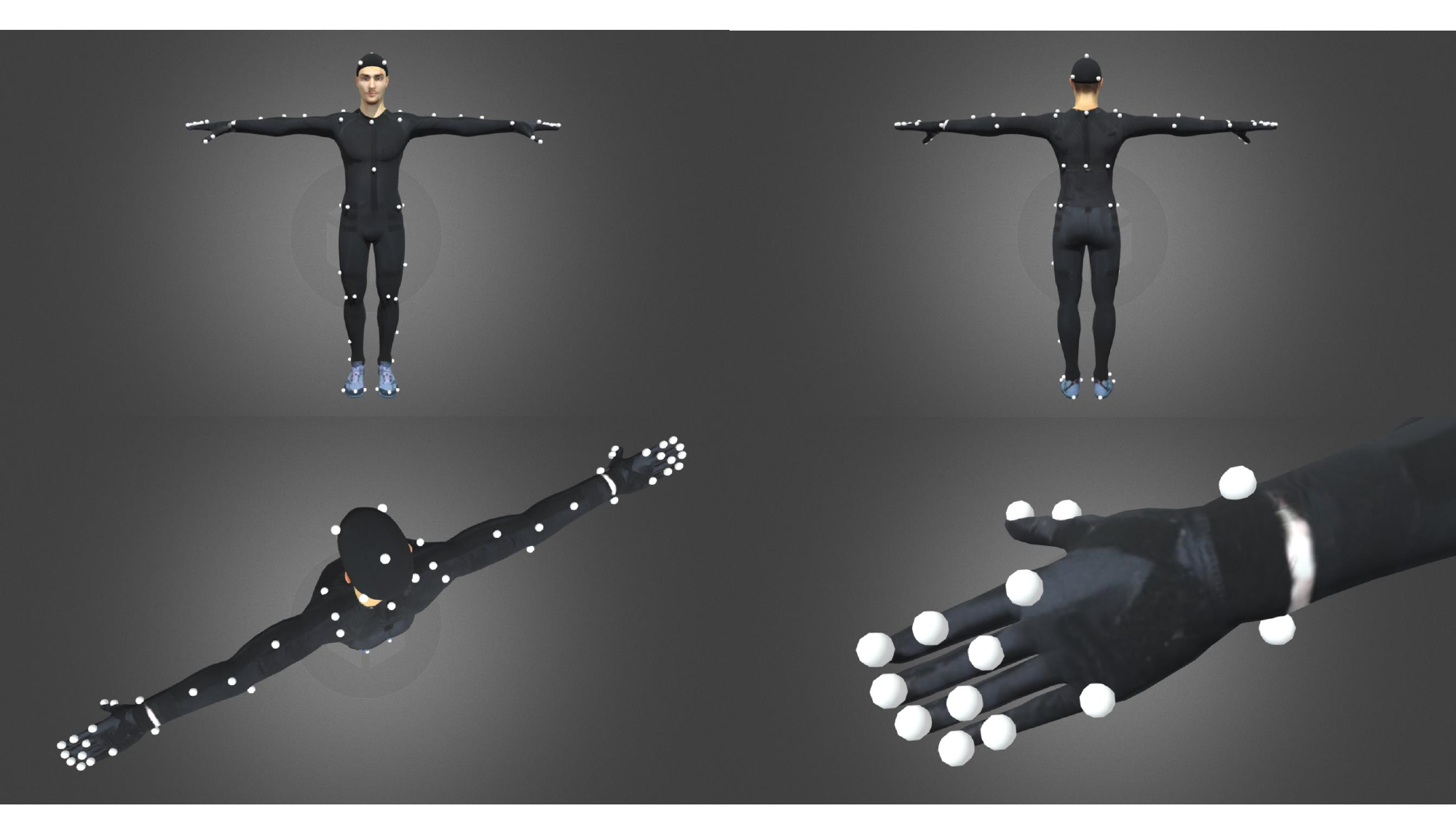}
    \caption{\textbf{Positions of motion capture markers.}}
    \label{fig:spfig3}
\end{figure}

\begin{figure}
    \centering
    \includegraphics[width=.7\textwidth]{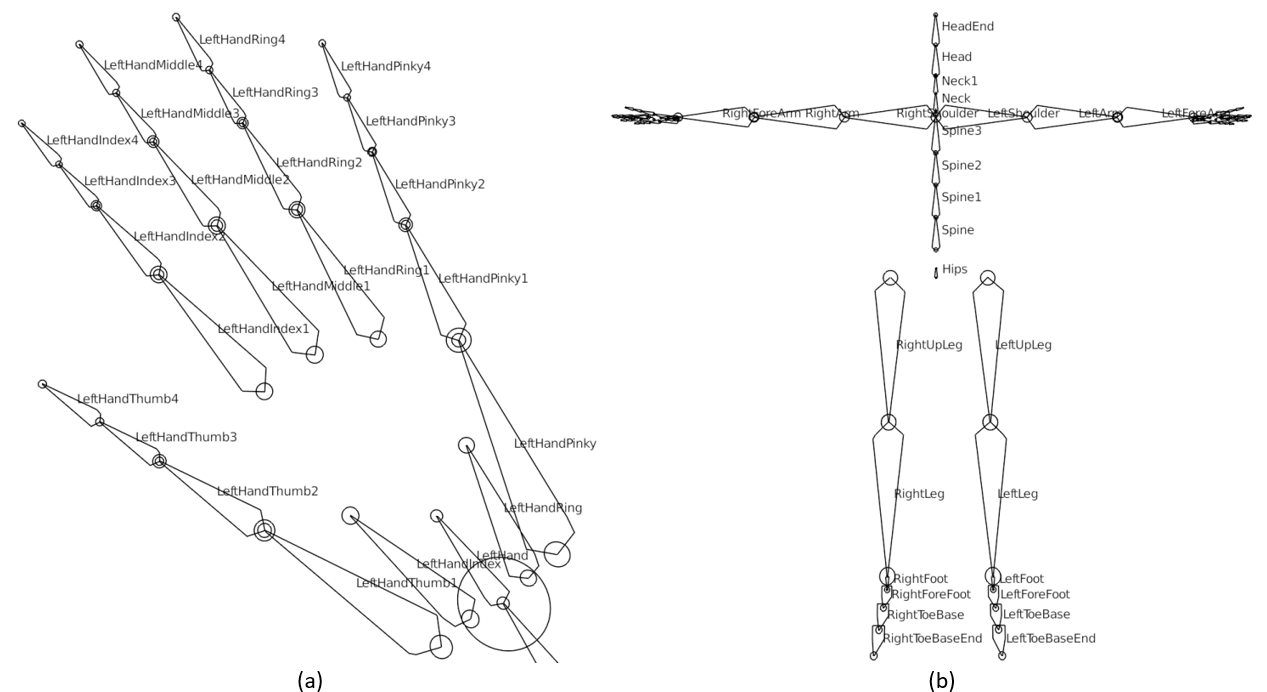}
    \caption{\textbf{Joint names for body and hands.} The
representative human skeleton has 48 hand joints (a) and 27 body joints (b).}
    \vspace{-0.5cm}
    \label{fig:spfig3}
\end{figure}

\begin{figure}
    \centering
    \includegraphics[width=.9\textwidth]{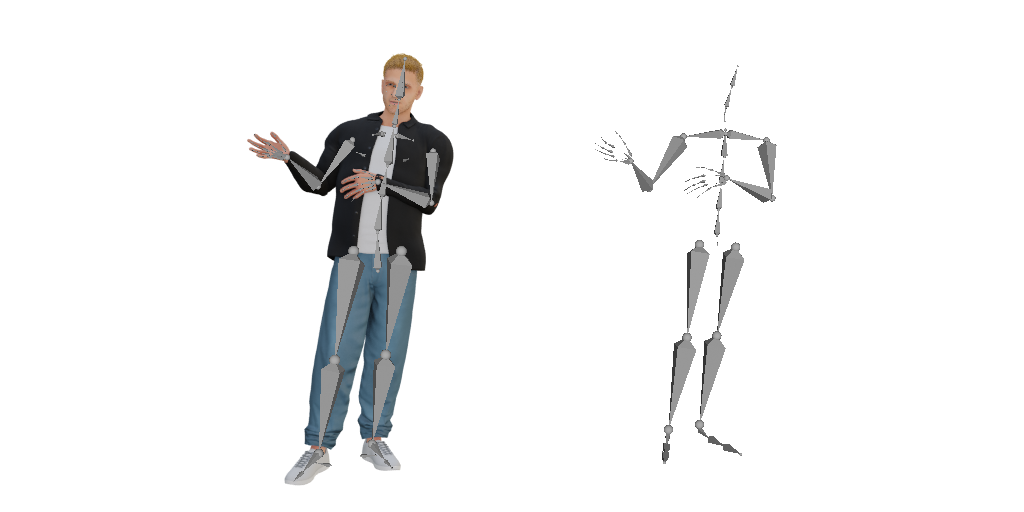}
    \caption{\textbf{Motion retargeting example: motion-driving skeleton and retargeted result.}}
    \label{fig:spfig5}
\end{figure}

\begin{figure}
    \centering
    \includegraphics[width=.9\textwidth]{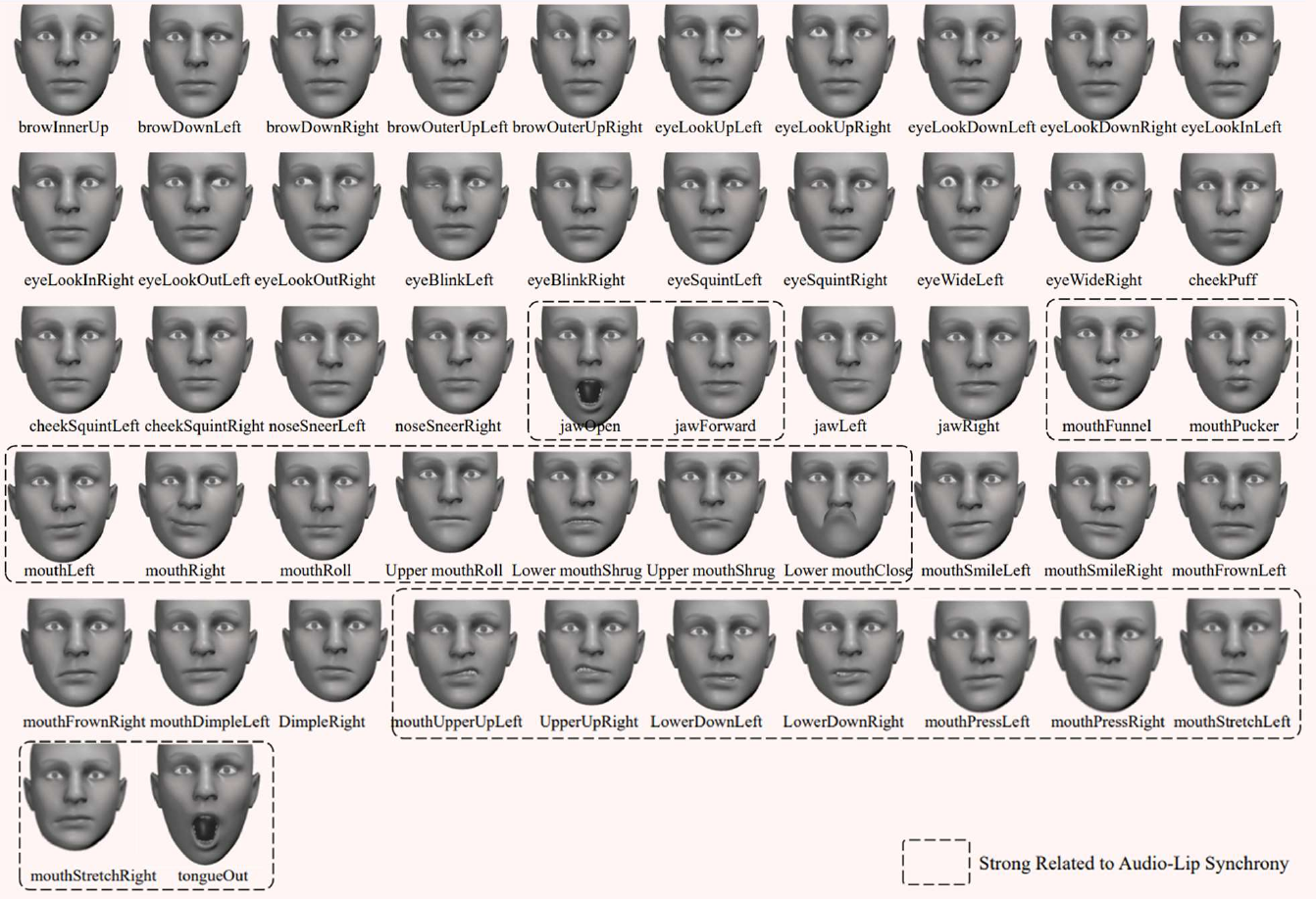}
    \caption{\textbf{The names for FACs (Facial Action Coding system) based blendshapes.}}
    \label{fig:spfig4}
\end{figure}

As shown in Figure \ref{fig:spfig4}, facial expressions are represented with the Facial Action Coding system (FACs) based blendshapes, where each expression only activates one part of the face (\textit{e.g.}, mouth area, eyes, eye-brows) at a time considering the human facial anatomy. 

All avatars used for demonstration were built by a Blender tool called HumanGeneratorv3\footnote{\url{https://www.humgen3d.com/}}.
We created avatars for 8 speakers. Furthermore, we also processed motion retargeting on the body bone animation. Thanks for giving the exception of license from the HumanGenerator team. These facial mesh data will be released together to better visualize the facial data recordings as an asset of the BEAT dataset.

\section{Additional Discussions for SRGR, FGD and BeatAlign.}
\begin{figure}
    \centering
    \includegraphics[width=0.8\textwidth]{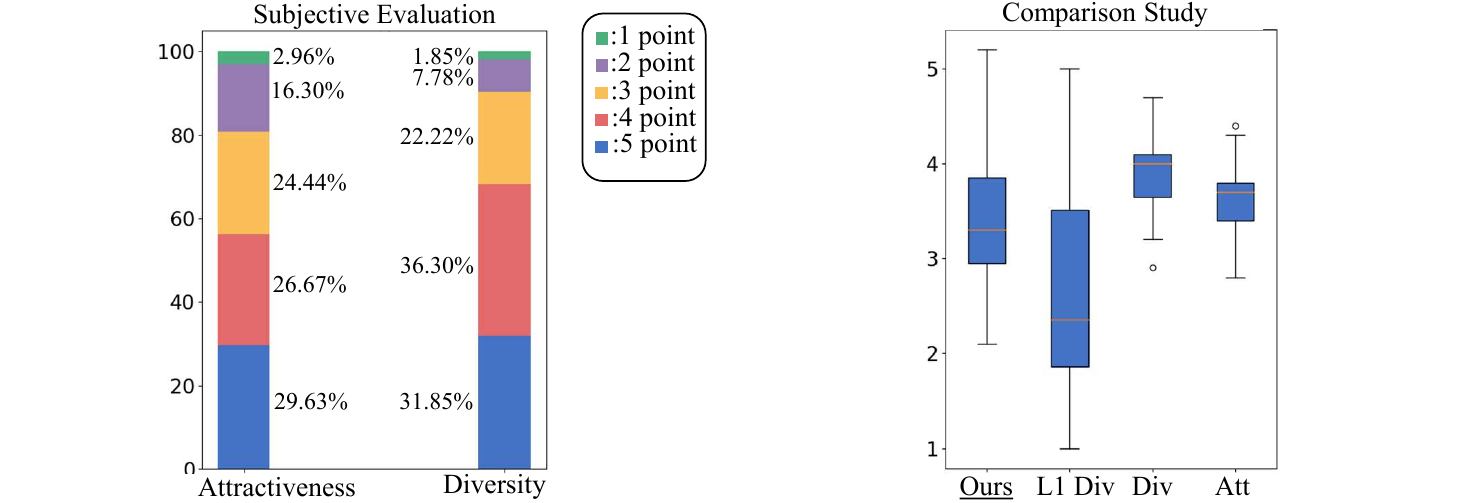}
    \caption{
\textbf{User Study Results for SRGR.} \textit{Left}: The proportional histogram shows the distribution of the score of 5-point Likert scale. \textit{Right}: The scores of comparison study illustrate that SRGR is more akin to subjective human perception in term of diversity and attractiveness when evaluating a gesture.
}
    \label{fig:spfig7}
\end{figure}

In the main paper, we demonstrate the SRGR is closer to human perception in the terms of diversity and attractiveness in comparison to the L1 Diversity in \cite{li2021audio2gestures}, considering the score distribution for each group of gesture clips. Here, we list the sum of scores distribution from all gesture clips in Figure \ref{fig:spfig7}. The experimental results are shown in Figure \ref{fig:spfig7} (left), which implies that there is a strong correlation between the attractiveness of a gesture and its diversity. More importantly, Figure \ref{fig:spfig7} (right) shows that SRGR is closer to the human perception in evaluating the diversity and attractiveness of gesture than the equal weight sum of L1 distance.

In addition, we also investigate two other metrics: Frechet Gesture Distance (FGD) \cite{yoon2020speech} and BeatAlign \cite{li2021ai}. We observed that they have few limitations for gesture synthesis evaluation. The calculation of FGD mainly depends on the gesture feature representation, but there is no common or well-defined gesture feature representation standard, due to different number of joints and different duration of analyzed segments. Besides, we found that some synthesized gesture sequences in the results received relatively constant FGD scores, however, there are obvious jitters that occurred in these gesture sequences. Although this problem might be solved by evaluating jointly with BeatAlign, it still suggests a better detection of physical correctness, or exploring a more generalized gesture feature extraction network.

The correctness of BeatAlign \cite{li2021ai} was verified on dance generation. Therefore, we propose to verify the feasibility and error rate of it on conversational gesture generation. Unlike dance generation, we extract the RMS onset of the audio as the audio beat, and for the gestures we take the local minimum of the velocity. The experimental results show that: i) for a random sample of 300 gestures clips, there is a 6\% higher score than GT; ii) for 100 gestures clips with 0.1 second steps and five seconds of panning back and forth, there is an average precision of 83\%; iii) Single directional evaluation has higher precision than the bi-directional evaluation (71\%), and non-exponential evaluation (59\%). Thus, although BeatAlign can be used as a metric to evaluate audio-gesture synchrony in conversational gesture generation, there is still room for improvement.

We also list formulas of L1 Div., FGD, and BeatAlign bellow for reference:
L1 diversity, which is the equal weight sum of L1 distance from different $N$ clips, as
\begin{equation}
\resizebox{.50\hsize}{!}{$
    L1 Div. =  \frac{1}{2 N (N-1)} \sum_{t=1}^{N} \sum_{j=1}^{N} \left\|p_{t}^{i}-\hat{p}_{t}^{j}\right\|_{1},$} 
\end{equation}

FGD\cite{yoon2020speech} is the FID calculated by a pretrained gesture encoder, as
\begin{equation}
\label{eqfid}
\resizebox{.60\hsize}{!}{$
\operatorname{FGD}(\textbf{m}, \hat{\textbf{m}})=\left\|\mu_{r}-\mu_{g}\right\|^{2}+\operatorname{Tr}\left(\Sigma_{r}+\Sigma_{g}-2\left(\Sigma_{r} \Sigma_{g}\right)^{1 / 2}\right),$}
\end{equation}
where $\mu_{r}$ and $\Sigma_{r}$ are the first and second moments of the latent features distribution $z_{r}$ of real human gestures \textbf{m}, and $\mu_{g}$ and $\Sigma_{g}$ are the first and second moment of the latent features distribution $z_{g}$ of generated gestures $\hat{\textbf{m}}$. BeatAlign \cite{li2021ai} is calculated as
\begin{equation}
\label{align}
\resizebox{.50\hsize}{!}{$
\text {BeatAlign}= \frac{1}{G} \sum_{b_{G}\in G} \exp \left(-\frac{\min _{b_{A}\in A}\left\|b_{G}-b_{A}\right\|^{2}}{2 \sigma^{2}}\right),$}
\end{equation}
Where $G, A$ is the set of gesture beat and audio beat, respectively. $\sigma$ is adjusted based on fps and we set 0.3 in our paper.

\section{More Subjective Results and Videos.}
\begin{figure*}[]

\begin{center}
\includegraphics[width=0.80\columnwidth]{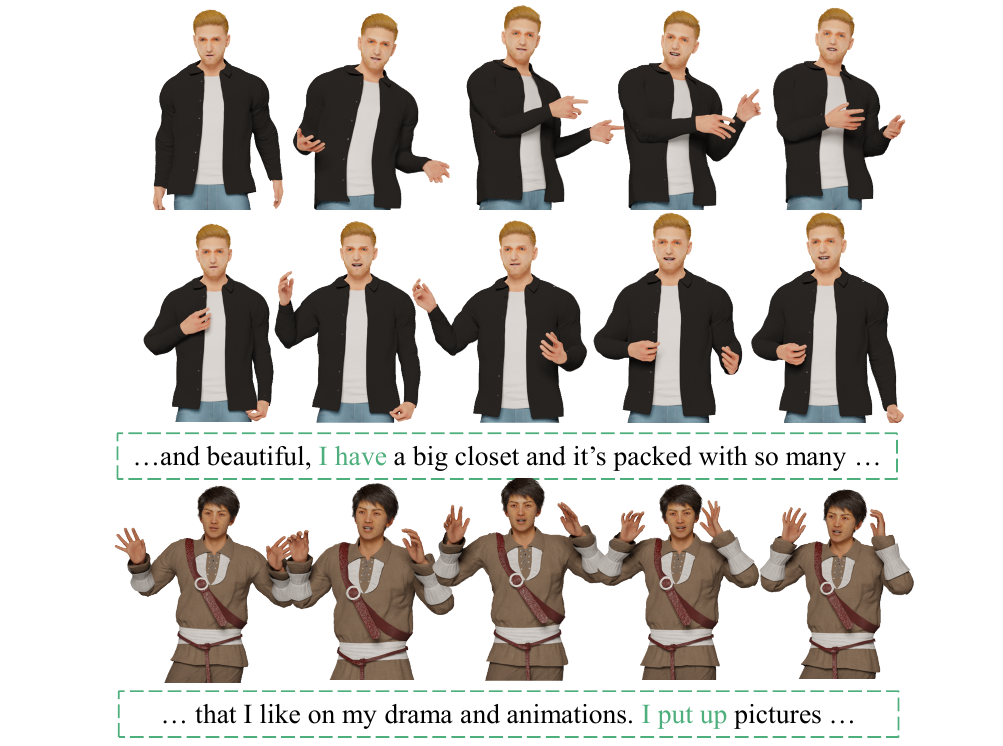}
\end{center}
\caption{
%
%
\textbf{Results Visualization.} Ground truth (top) and generated results with neutral (middle) and fear (down) emotions. 
}
\label{fig:fig7}
\end{figure*}

The subjective results of generated gestures are shown in Figure \ref{fig:fig7}. More video results and data are available on the project page. We also provide results on other languages, \textit{e.g.}, Japanese data.

\section{Details of baseline training}
Currently, we do not split the dataset based on speaker, \textit{i.e.}, some speakers only exist in the validation/test data, since the speaker ID is one of the inputs. For each speaker, we use the ratio 10:1:1 for the train/valid/test data splits. For baseline training, we select the best model based on the lowest validation FGD score during training for all baseline models. The final selected epoch is listed in Table \ref{tab:sptab1}. 
\begin{table}
\centering
\caption{\textbf{Best Epoch for Baselines.}}
\label{tab:sptab1}
\begin{tabular}{lccccc}
\multicolumn{1}{c}{} & Seq2Seq & S2G & A2G & MultiContext & Ours (CaMN)  \\ 
\hline
FGD                  & 261.3   &256.7     &223.8     &176.2         &123.7             \\
Learning Rate                  &1e-3         &1e-4     &1e-4     &5e-4              &2e-4              \\
Epoch                &103         &87     &271     &129             & 117            
\end{tabular}
\end{table}

\end{document}